\PassOptionsToPackage{numbers,sort&compress}{natbib} 

\documentclass[preprint,a4paper]{elsarticle}

\usepackage{amssymb}
\usepackage{amsthm}

\usepackage{lineno}

\usepackage{amsmath,nccmath}
\usepackage{float}
\usepackage{subfig}
\usepackage{mathtools}
\usepackage{multirow}

\newcommand\mystrut{\rule{-2pt}{15pt}}

\journal{Journal of Sound and Vibration}

\begin{document}

\begin{frontmatter}

\title{Modelling variability in vibration-based PBSHM via a generalised population form}

%

\author[add1]{T.A.\ Dardeno\corref{mycorrespondingauthor}}
\cortext[mycorrespondingauthor]{Corresponding author}
\ead{t.a.dardeno@sheffield.ac.uk}

\author[add2]{L.A.\ Bull}

\author[add1]{R.S.\ Mills}

\author[add1]{N.\ Dervilis}

\author[add1]{K.\ Worden}

\address[add1]{Dynamics Research Group, Department of Mechanical Engineering, \\ University of Sheffield, Sheffield S1 3JD, UK}

\address[add2]{The Alan Turing Institute, The British Library, London NW1 2DB, UK}

\begin{abstract}
	Structural health monitoring (SHM) has been an active research area for the last three decades, and has accumulated a number of critical advances over that period, as can be seen in the literature. However, SHM is still facing challenges because of the paucity of damage-state data, operational and environmental fluctuations, repeatability issues, and changes in boundary conditions. These issues present as inconsistencies in the captured features and can have a huge impact on the practical implementation, but more critically, on the generalisation of the technology. Population-based SHM has been designed to address some of these concerns by modelling and transferring missing information using data collected from groups of similar structures.

	In this work, vibration data were collected from four healthy, nominally-identical, full-scale composite helicopter blades. Manufacturing differences (e.g., slight differences in geometry and/or material properties), among the blades presented as variability in their structural dynamics, which can be very problematic for SHM based on machine learning from vibration data. This work aims to address this variability by defining a general model for the frequency response functions of the blades, called a \emph{form}, using mixtures of Gaussian processes.

\end{abstract}




\end{frontmatter}


\section{Introduction}
\label{}
	
	Engineering structures are inherently uncertain, even when structures can be considered \emph{nominally-identical}, because of manufacturing differences causing variation in geometry and material properties. These differences, as well as variations caused by ageing parts and changes in testing conditions (e.g., operational and environmental fluctuations or changes in boundary stiffness), make generalisation among structures difficult, particularly with respect to their dynamic properties \cite{Dowsett,CHEN200664}. The current work is focussed on applying structural health monitoring (SHM), to a set of nominally-identical (i.e., \emph{homogeneous} \cite{Bull_1,Bull_2}) structures for the purpose of damage detection. (As opposed to a \emph{heterogeneous} population \cite{gosliga2021foundations,gardner2021foundations}, where the members are more disparate, such as different designs of a suspension bridge). However, these fluctuations unrelated to damage are challenging for SHM, as damage-sensitive features may also be sensitive to normal variations, which may cause issues with discerning between damaged and healthy states \cite{Worden2002NoveltyDI,Alampalli,Cawley,HoonSohn}. For example, it is well-known that structural damage is often associated with a reduction in stiffness. Depending on the location and severity of the damage, and the complexity of the examined mode, this damage may present as a decrease in natural frequency. Increased ambient temperature, loosening of bolts at a support, or similar situations may also reduce stiffness and can mimic damage. Likewise, conditions that increase stiffness (e.g., decreased ambient temperature) can increase natural frequency and mask damage.
	
	This sensitivity of natural frequency to environmental/operational changes in addition to damage has been demonstrated experimentally \cite{Alampalli,Cawley,HoonSohn}. Alampalli \cite{Alampalli}, performed impact modal testing on a small 6.76 x 5.26 metre bridge, before and after damaging the bridge with a saw cut made across the bottom flanges of both the main girders. They found that for temperatures above freezing, the stiffness reduction caused by damage was reflected as a downward shift in natural frequencies (-2\% to -11\%, depending on the mode), as expected. However, they also found that for temperatures below freezing, the measured frequencies increased significantly, primarily from freezing of accumulated moisture within the supports (+25\% to +58\%, depending on the mode), which masked the reduction caused by damage. In addition, uncertainty among nominally-identical structures (e.g., slight variation in internal structure and material properties), like those caused by environmental and other fluctuations, may be erroneously flagged as damage or mask subtle damage in a structure. Cawley \cite{Cawley}, initiated a crack at the fixed end of a cantilever beam and evaluated the natural frequencies as the beam length varied. He found that the change in natural frequency resulting from a cut through 2\% of the beam depth was 40 times smaller than that caused by a 2\% increase in the beam length \cite{Cawley,HoonSohn}. Given that these small differences among structures and testing conditions can have such a large impact on dynamics and damage detection, one would ideally aim to collect new data for each structure under each set of testing conditions. However, collecting comprehensive data covering a range of operational and damaged conditions from each individual structure is typically not feasible (a wind farm, for example, may contain hundreds or even thousands of wind turbines). In such cases, population-based SHM (PBSHM) can be used to make inferences between different members of a \emph{population}, i.e., groups of similar systems \cite{Bull_1,Bull_2,gosliga2021foundations,gardner2021foundations,Bull_3}. 

	PBSHM seeks to transfer valuable information across similar structures. In certain cases, the behaviour of the group can be represented using a general model. Bull \emph{et al.}\ used Gaussian process (GP) regression of real \cite{Bull_1,Bull_2}, and imaginary \cite{Bull_2}, frequency response functions (FRFs) to develop a generic representation, called a \emph{form}, of a population of nominally-identical, but slightly different, eight degree-of-freedom (DOF) systems. The population was comprised of multiple simulated healthy and damaged systems \cite{Bull_1,Bull_2}, and one experimental rig, which was used to test the form in \cite{Bull_1}. Normal-condition data from several members of the population were used to train the form, and the posterior predictive distributions of the GP were used to aid novelty detection across the population, using the Mahalanobis squared-distance novelty index \cite{WORDEN2000647}. The form was found to predict accurately the healthy and damaged states across the population \cite{Bull_1,Bull_2}. L\'azaro-Gredilla \emph{et al.}\ \cite{LAZAROGREDILLA20121386}, introduced the overlapping mixture of Gaussian processes (OMGP), which uses a variational Bayesian approach to learn hyperparameters and labels by clustering data according to various trajectories across the input space. This technique was applied to the population form in \cite{Bull_1,Bull_3}, using an OMGP to infer multivalued wind-turbine power-curve data, with categorisation of the data performed in an unsupervised manner. Novelty detection was performed using the negative log-marginal-likelihood of the OMGP, and correctly identified \emph{good} and \emph{bad} power curves for the majority of data tested \cite{Bull_1,Bull_3}.
			
	In this work, data were collected from four healthy, nominally-identical helicopter blades. FRFs were computed from the measured time-domain data, and these data were used to develop a form for the blades, first using a supervised mixture of GPs, and second using an (unsupervised) OMGP. The posterior predictive distributions from the OMGP were later applied to a novelty detection approach to evaluate new (simulated) FRF data against the form. As such, this work builds upon that performed in \cite{Bull_1,Bull_2}, by extending the concept of a population form to a set of nominally-identical helicopter blades with a range of undamaged-condition FRF data, using mixtures of probabilistic regression models \cite{Bull_1,Bull_3,LAZAROGREDILLA20121386}, to account for the large variability in the FRFs of the blades. The significant discrepancies in the experimental data set are representative of the differences in the blades in practice, with variability arising from small variations in the internal structure and material properties of the helicopter blades, as well as slight changes in boundary conditions. The small population of helicopter blades provided a useful data set to develop and test the techniques presented herein, such that these methods can be later applied to other homogeneous populations such as a wind farm. 
		
	Although the specific application of this work relates to population-based damage detection, this work is also applicable to the more general problem of dynamic analysis of a population. Attempts to quantify uncertainty in vibration data have included parametric \cite{HONG20111091}, and nonparametric \cite{CHEN200664,Yang}, approaches. Parametric models allow incorporation of engineering knowledge, but can suffer from over-training (if highly complex), or over-simplification. Nonparametric models, such as the approach used in \cite{CHEN200664}, and the polynominal-chaos expansion used in \cite{Yang}, in general do not incorporate prior engineering knowledge, and this can lead to problematic solutions because of the high model flexibility. Gaussian processes provide a good middle ground, as they are nonparametric, but allow for the use of a parametric mean function where prior knowledge about the data can be incorporated into the model. The (OM)GP form provides a probabilistic model of the FRF that can capture uncertainty caused by the variations described above as well as measurement noise.
	
	This work formalises the use of a modal analysis approach to represent a group of similar \emph{dynamic} systems, using a data-driven model (or representation) referred to as the \emph{population form} \cite{Bull_1,Bull_2}. Unlike previous work \cite{Bull_2}, the form here uses a mixture of regressions to represent \emph{population dynamics/FRFs} - rather than a single, averaged model (complete pooling). In turn, the modelling procedure can approximate population data with more complex variations and different modes in the posterior distribution; for example, relating to distinct operating conditions, characteristic subgroups in the population, or individual members. While mixture models have been used to represent the \emph{form} in previous work, relating to wind farm power prediction \cite{Bull_1,Bull_3}, these models were not physics informed \cite{Cross2022}; here, in contrast, the proposed (semi-parametric) form is constructed around the established theory of linear modal analysis – allowing for the natural inclusion of domain (engineering) expertise via interpretable parameters (e.g., natural frequencies, damping ratios, mode shapes).

	The layout of this paper is as follows. Section 2 provides an overview of the theory used in this work, including modal analysis, conventional GPs, and OMGP. Section 3 describes the experimental campaign performed on the population of nominally-identical helicopter blades and introduces the data set used in the analyses. Section 4 discusses how a generalised normal condition was determined for the helicopter blades, first using a supervised mixture of GPs, and second using an (unsupervised) OMGP. Finally, Section 5 demonstrates how the OMGP can be used to inform damage detection, by computing the marginal likelihood for a series of (simulated) damaged FRFs.
	
\section{Background theory}
\label{}
	
	In this work, a generalised normal condition was determined for a small population of nominally-identical helicopter blades. A form was developed for the population, first using a supervised mixture of GPs, and second using an (unsupervised) OMGP. A modal-based FRF estimation technique was parameterised and used to compute a non-zero prior mean function for the GPs. The marginal likelihood (evidence) was then used along with the posterior predictive distributions of the OMGP to evaluate (simulated) new data for novelty. A theoretical background for these methodologies is provided below. 
	
	\subsection{Modal analysis}
	
	The equation of motion for a multiple DOF system can be written as,

	\begin{equation}
		\mathbf{M}\ddot{\mathbf{u}}(t) + \mathbf{C}\dot{\mathbf{u}}(t) + \mathbf{K}{\mathbf{u}}(t) = \mathbf{z}(t)
		\label{eq:EOM}
	\end{equation}

	\noindent where $ \ddot{\mathbf{u}}(t) $, $ \dot{\mathbf{u}}(t) $, $ {\mathbf{u}}(t) $, and $ \mathbf{z}(t) $ are acceleration, velocity, displacement, and force, respectively. In most cases, the mass $ \mathbf{M} $, damping $ \mathbf{C} $, and stiffness $ \mathbf{K} $ matrices are coupled. For linear systems, and in the absence of viscous damping, the equation of motion can be decoupled, such that the system is represented by multiple single degree-of-freedom (SDOF) oscillators. This decoupling is performed via the eigenvalue expression,

	\begin{equation}
		[\mathbf{K}-\omega_n^2\mathbf{M}]\boldsymbol{\Psi} = 0
		\label{eq:eig}
	\end{equation}

	\noindent and yields the natural frequencies in radians, $ \omega_n $, and mode shapes, $ \boldsymbol{\Psi} $, of the system. 
	
	The physical equation of motion can then be cast in modal space to give the uncoupled modal equations, with modal coordinates $ {\mathbf{p}}(t) $, written as,
	
	\begin{equation}
	 	\boldsymbol{\Psi}^\text{T}\mathbf{M}\boldsymbol{\Psi}\ddot{\mathbf{p}}(t)+ \boldsymbol{\Psi}^\text{T}\mathbf{C}\boldsymbol{\Psi}\dot{\mathbf{p}}(t) + \boldsymbol{\Psi}^\text{T}\mathbf{K}\boldsymbol{\Psi}{\mathbf{p}}(t) = \boldsymbol{\Psi}^\text{T}\mathbf{z}(t)
		\label{eq:modalEOM}
	\end{equation}
	
	\noindent where,
	
	\begin{equation}
		\mathbf{u}(t) = \boldsymbol{\Psi}{\mathbf{p}}(t)
		\label{eq:phystomodal}
	\end{equation}
	
	Frequency response functions (FRFs) can be used to visualise the natural frequency components of a system, and are computed by normalising the response signal at a given location to the excitation force. This work used the $ \text{H}_1 $ estimator to compute FRFs. For a response at location $ i $ resulting from excitation at $ j $, the $ \text{H}_1 $ estimator is computed as,
	
	\begin{equation}
		\mathbf{H}_{ij}(\omega) = \frac{\mathbf{G}_{zu}(\omega)}{\mathbf{G}_{zz}(\omega)}
		\label{eq:FRF}
	\end{equation}
	
	\noindent where,
	
	\begin{equation}
		\begin{split}
			& \mathbf{G}_{zu}(\omega) \overset{\Delta}{=} {\it \mathbb{E}} [\mathbf{S}_{u_i}(\omega) \mathbf{S}_{z_j}^*(\omega)] \\
			& \mathbf{G}_{zz}(\omega) \overset{\Delta}{=} {\it \mathbb{E}} [\mathbf{S}_{z_j}(\omega) \mathbf{S}_{z_j}^*(\omega)]
		\end{split}
		\label{eq:spectra}
	\end{equation}
	
	\noindent and,

	\begin{equation}
		\begin{split}
			& \mathbf{S}_{z_j}(\omega) \overset{\Delta}{=} \mathcal{F}[{\mathbf{z}_j}(t)] \\
			& \mathbf{S}_{u_i}(\omega) \overset{\Delta}{=} \mathcal{F}[{\mathbf{u}_i}(t)]
		\end{split}
		\label{eq:FFT}
	\end{equation}
	
	\noindent The asterisk $ * $ denotes complex conjugation, $ \omega $ is frequency in radians, and $ \mathcal{F} $ is a discrete Fourier transform (this work used a fast Fourier transform or FFT). The input force in the time domain at location $ j $ is $ {\mathbf{z}_j}(t) $, and the output response (i.e., acceleration, velocity, or displacement) in the time domain at $ i $ is $ {\mathbf{u}_i}(t) $.
	
	With assumed linear behaviour and proportional damping, the accelerance FRF (i.e., given acceleration response data) can also be estimated using modal parameters,
	
	\begin{equation}
		\mathbf{H}_{ij}(\omega) = -\omega^2 \sum_{k=1}^{n} \frac{\text{A}_{ij}^{(k)}}{\omega_{nk}^2-\omega^2+2i\zeta_k\omega\omega_{nk}}
		\label{eq:modalFRF}
	\end{equation}
	
	\noindent where $ \text{A}_{ij}^{(k)} $ is the residue for mode $ k $, defined as the product of the mass-normalised mode shapes at locations $ i $ and $ j $ ($ \text{A}_{ij}^{(k)} = \psi_{ik}\psi_{jk} $) \cite{wordennonlinearity}. The natural frequency associated with mode $ k $ is $ \omega_{nk} $, and the modal damping associated with mode $ k $ is $ \zeta_k $ \cite{wordennonlinearity}. In the examples below, Eq.\ (\ref{eq:modalFRF}) was used to compute a non-zero prior mean function for the (OM)GP form.

	\subsection{Conventional Gaussian process regression}

	Gaussian processes provide a Bayesian nonparametric approach to regression/classification and are a suitable basis for a population form because of their robust uncertainty quantification. For a set of training data $ \mathcal{D} = \{x_i,y_i\}_{i=1}^N = \{{\mathbf{x},\mathbf{y}}\} $, where $ x_i $ and $ y_i $ are sets of $ N $  inputs and outputs, respectively, the GP regression attempts to obtain the predictive distribution for observation $ y_* $ given a new input $ x_* $ and training data  $ \mathcal{D} $ \cite{LAZAROGREDILLA20121386}. The GP regression model can be expressed as a noiseless latent function $ f(x_i) $ plus an  independent noise term $ \epsilon_i $ \cite{LAZAROGREDILLA20121386}, written as,
	
	\begin{equation}
		y_i = f(x_i) + \epsilon_i \ 
		\label{eq:latentfunc}
	\end{equation}

	A GP prior is placed over the latent function $ f(x_i) $ and a Gaussian prior is placed over the noise term $ \epsilon_i $,

	\begin{equation}
		f(x_i) \sim \mathcal{GP}(m(x_i),k(x_i,x_j)), \; \epsilon_i \sim \mathcal{N}(0,\sigma^2)  \
		\label{eq:GP}
	\end{equation}

	\noindent where $ m(x_i) $ is a mean function, $ k(x_i,x_j) $ is a covariance function, and $ \sigma^2 $ is a noise variance term and the first hyperparameter. Over a finite, abitrary set of inputs $ \mathbf{x} = \{x_1,...,x_N\} $, the GP is a (joint) multivariate Gaussian \cite{Bull_3,murphy2012machine}, 

	\begin{equation}
		p\left(\mathbf{f} \;|\; \mathbf{x}\right) = \mathcal{N}\left(\mathbf{m},\mathbf{K_{xx}}\right)
		\label{eq:GP2}
	\end{equation}

	\noindent where $ \mathbf{m} = \{m(x_i),...,m(x_N)\} $, $ \mathbf{f} = \{f(x_i),...,f(x_N)\} $, and $ \mathbf{K_{xx}} $ is the covariance matrix, such that $ \mathbf{K_{xx}}[i,j] = k(x_i,x_j) \; \forall i,j \in \{1,...,N\} $ \cite{Bull_3}. Note that in certain instances within this paper, square brackets are used to index matrices and vectors for readability. 

	The mean function $ m(x_i) $ is often assumed to be zero, given that the GP is flexible enough to model arbitrary trends  \cite{murphy2012machine,3569,Bull_3}. However, the GP fit may be improved, by incorporating \emph{a priori} knowledge from engineering judgement or an understanding of the system physics, into the GP prior by choosing suitable mean and covariance functions \cite{Bull_1,Bull_3}. Coupling behaviour between $ y_i $ and $ y_j $, and therefore important properties such as process variance and smoothness, is specified by the covariance function \cite{LAZAROGREDILLA20121386}. Appropriate covariance functions are selected according to the shape and smoothness of the data. This work applied the commonly-used squared exponential,

	\begin{equation}
		k(x_i,x_j) = \sigma_f^2 \exp 	\left(-\frac{1}{2l^2}(x_i-x_j)^2\right)  \
		\label{eq:sq_exp}
	\end{equation}

	\noindent where $ \sigma_f^2 $ and $ l $ are second and third hyperparameters, representing the process variance and length scale, respectively. The length scale $ l $ influences how fast the correlation between outputs decays according to the input separation, and the process variance $ \sigma_f^2 $ determines the variance of the modelled signal \cite{LAZAROGREDILLA20121386}.

	The remaining hyperparameters {\boldmath $\alpha$} are those required by the mean function, if non-zero, giving the full model hyperparameters as $ \boldsymbol{\theta} = \{l, \sigma_f, \sigma, \boldsymbol{\alpha}\} $. In this work, the modal damping-based FRF estimation from Eq.\ (\ref{eq:modalFRF}) provided the mean function for the GP regression. Modal damping ($ \zeta_k $), residue ($ \text{A}_{ij}^{(k)} $), and natural frequency ($ \omega_{nk} $), provided the mean-function hyperparameters. Keeping $ \boldsymbol{\theta} $ fixed, the joint distribution of the training data $ \mathcal{D} = \{{\mathbf{x},\mathbf{y}}\} = \{x_i,y_i\}_{i=1}^N $ and unknown function values at test locations under the prior $\{{\mathbf{x_*},\mathbf{y_*}}\} = \{\mathbf{x_*}[i],\mathbf{y_*}[i]\}_{i=1}^M $ is a multivariate Gaussian, written as, 
	
	\begin{gather}
		\begin{bmatrix*}[l] 
			\mathbf{y} \\
			\mathbf{y_*} 
		\end{bmatrix*}
		\sim 
		\mathcal{N}
		\left(
		\begin{bmatrix*}[l]	
			\mathbf{m} \\ 
			\mathbf{m_*} 
		\end{bmatrix*},
		\begin{bmatrix*}[c]	
			\mathbf{K_{xx}} + \mathbf{R} &
			\mathbf{K_{xx_*}}                \\ 
			\mathbf{K_{x_*x}}                 &
			\mathbf{K_{x_*x_*}} + \mathbf{R_*}
		\end{bmatrix*}
		\right)
		\label{eq:multivarGauss}
	\end{gather}

	\begin{equation}
	\begin{split}
		& \mathbf{R} \overset{\Delta}{=} \sigma^2\mathbf{I}_N  \\
		& \mathbf{R_*} \overset{\Delta}{=} \sigma^2\mathbf{I}_M 
	\end{split}
	\label{eq:defineR}
	\end{equation}

	\noindent where $ \mathbf{I} $ is an identity matrix of dimension $ N $ or $ M $ as specified \cite{Bull_3,3569}. Likewise, $ \mathbf{m_*} = \{m\left(\mathbf{x_*}[i]\right)\}^M_{i=1} $ is the mean vector for new observations \cite{Bull_3}.

	Conditioning the joint distribution in Eq.\ (\ref{eq:multivarGauss}) on the training data in $ \mathcal{D} $ yields the predictive distribution over $ \mathbf{y_*} $, with mean $ \boldsymbol{\mu_*} $ and covariance $ \boldsymbol{\varSigma_*} $ \cite{3569}, defined as,
	
	\begin{equation}
		\begin{split}
			& p(\mathbf{y_*} \;|\; \mathbf{x_*}, \mathcal{D}) = \mathcal{N}(\boldsymbol{\mu_*},\boldsymbol{\varSigma_*}) \\
			& \boldsymbol{\mu_*} \overset{\Delta}{=} \mathbf{m_*} + \mathbf{K_{x_*x}}\left(\mathbf{K_{xx}} + \mathbf{R}\right)^{-1} \left(\mathbf{y}-\mathbf{m}\right) \\
			& \boldsymbol{\varSigma_*} \overset{\Delta}{=} \mathbf{K_{x_*x_*}} - \mathbf{K_{x_*x}}\left(\mathbf{K_{xx}} + \mathbf{R}\right)^{-1}\mathbf{K_{xx_*}} + \mathbf{R_*}
		\end{split}
		\label{eq:GPposterior}
	\end{equation}

	This work applied a Type-II maximum likelihood approach to learn the hyperparameters, which maximises the marginal likelihood of the model $ p(\mathbf{y} \;|\; \mathbf{x}, \boldsymbol{\theta}) $ \cite{3569}. This procedure involves minimising the negative log-marginal-likelihood to improve numerical stability \cite{murphy2012machine}. Therefore, the hyperparameters were optimised via,

	\begin{equation}
		\hat{\theta} = \underset{\theta}{\operatorname{argmin}} 	\{-\log p(\mathbf{y} \;|\; \mathbf{x}, \boldsymbol{\theta})\}
		\label{eq:neglikelihood}
	\end{equation}

	\noindent where,

	\begin{equation}
		\begin{split}
			& -\log p(\mathbf{y} \;|\; \mathbf{x}, 	\boldsymbol{\theta}) = -\log \mathcal{N} (\mathbf{y} \;|\; \mathbf{m}, \; \mathbf{K_{xx}} + \mathbf{R})  \\ 
			& = \frac{N}{2}\log(2\pi) + \frac{1}{2}\log|\mathbf{K_{xx}} + \mathbf{R}| \\
			& + \frac{1}{2}[(\mathbf{y}-\mathbf{m})^\textsf{T}(\mathbf{K_{xx}} + \mathbf{R})^{-1}(\mathbf{y}-\mathbf{m})]
		\end{split}
		\label{eq:optimisation}
	\end{equation}

	Constrained nonlinear optimisation was performed via the \emph{fmincon} function in MATLAB, and appropriate bounds were applied to the mean-function hyperparameters, given prior knowledge of the modal characteristics of the helicopter blades.

	\subsection{An overlapping mixture of Gaussian processes (OMGP)}
	In one example, this work used the conventional GP equations to evaluate a supervised mixture of GPs by assuming known which data belonged to each helicopter blade, and fitting each data set separately. In a second example, the data were fitted using an overlapping mixture of Gaussian processes (OMGP), where the number of trajectory functions were assumed known \emph{a priori}, but it was not assumed which data belonged to a given class. In this case, the unknown class was the particular helicopter blade that produced the measurements. In practice, it is likely that this information would be known; however, there are regularly other forms of missing data labels, such as changes in operating conditions, or possible damage. The helicopter blades case study is also representative of these types of situations, commonly encountered in PBSHM, where certain information is unavailable. As it is an unsupervised learning algorithm, the OMGP is a useful and flexible method for developing the population form in the absence of complete information.

	The OMGP \cite{LAZAROGREDILLA20121386,tay2008modelling,Bull_1,Bull_3}, describes a data set by evaluating each observation using one of \emph{K} latent functions plus additive noise,
	
	\begin{equation}
		y_i^{(k)} = \{f^{(k)}(x_i) + \epsilon_i\}_{k=1}^{K} 
		\label{eq:latentfuncOMGP}
	\end{equation}

	The OMGP is unsupervised, i.e., the labels that assign the data to a given function are unknown. Therefore, another latent variable is introduced, $ \mathbf{Z} $, which is a binary indicator matrix. The matrix $ \mathbf{Z} $, with entries $ \mathbf{Z}[i,k] $, is entirely populated with zeroes, except for one non-zero entry per row. The non-zero entries, $ \mathbf{Z}[i,k] \neq 0 $, indicate that observation $ i $ was generated by function $ k $. The likelihood of the OMGP is therefore written as \cite{LAZAROGREDILLA20121386,Bull_3}, 

	\begin{equation}
		p(\mathbf{y} \;|\; \{\mathbf{f}^{(k)}\}_{k=1}^{K}, \mathbf{Z}, \mathbf{x}) =
		\prod_{i,k = 1}^{N,K}
		 p{(y_i \;|\; {f}^{(k)}(x_i))}^{\mathbf{Z}[i,k]}
		\label{eq:likelihoodOMGP}
	\end{equation}

	Prior distributions are then placed over the latent functions and variables, such that \cite{Bull_3},
	
	\begin{equation}
		P(\mathbf{Z}) = \prod_{i,k = 1}^{N,K}  	{\boldsymbol{\varPi}[i,k]}^{\mathbf{Z}[i,k]} 
	\label{eq:priorOMGP1}
	\end{equation}

	\begin{equation}
		f^{(k)}(x_i) \sim \mathcal{GP}(m^{(k)}(x_i),k^{(k)}(x_i,x_j)), \epsilon_i \sim \mathcal{N}(0,\sigma^2)  
	\label{eq:priorOMGP2}
	\end{equation}	

	\noindent where $ P(\mathbf{Z}) $ is the prior over the indicator matrix $ \mathbf{Z} $, and $ \boldsymbol{\varPi}[i,:] $ is a histogram over the $ K $ components for the $ i $th observation, and $ \sum_{k=1}^{K}\boldsymbol{\varPi}[i,k] = 1 $ \cite{Bull_3}. The terms in Eq.\ (\ref{eq:priorOMGP2}) are independent GP priors over each latent function $ f^{(k)} $ with distinct mean $ m^{(k)}(x_i) $ and kernel $ k^{(k)}(x_i,x_j) $ functions \cite{Bull_3}. A shared hyperparameter $ \sigma $, is used to define the prior over the noise variances, to reduce the number of latent variables in the calculations \cite{Bull_3}. As with the conventional GP, this work used the modal damping-based FRF estimation from Eq.\ (\ref{eq:modalFRF}) as the mean function for the GP regression. Modal damping, residue, and natural frequency provided the mean-function hyperparameters. 
	
	\subsubsection{Variational inference}
	
	Exact computation of the posterior distribution $ p(\{\mathbf{f}^{(k)}\}_{k=1}^{K},\mathbf{Z} \;|\; \mathbf{x},\mathbf{y}) $ is intractable; therefore, a variational inference \cite{VIreview} and Expectation Maximisation (EM) scheme was employed to estimate the posterior, by specifying an approximate density family $ q(\mathbf{a}) \in \mathcal{Q} $ over the target conditional $ p(\mathbf{a} \;|\; \mathbf{b}) $ \cite{VIreview,LAZAROGREDILLA20121386,Bull_1,Bull_3}. The best approximation $ \hat{q}(\mathbf{a}) $ of the true posterior distribution $ p(\mathbf{a} \;|\; \mathbf{b}) $ can be expressed via the KL-divergence \cite{VIreview,Bull_3},

	\begin{equation}
		\hat{q}(\mathbf{a}) = \underset{q(\mathbf{a}) \in \mathcal{Q}}{\operatorname{argmin}} \{\text{KL}(q(\mathbf{a}) \;||\; p(\mathbf{a} \;|\; \mathbf{b}))\}
	\label{eq:KLdivergence1}
	\end{equation}	
	
    \begin{equation}
    	\begin{split}
		\text{KL}(q(\mathbf{a}) \;||\; p(\mathbf{a} \;|\; \mathbf{b})) 
		& = \mathbb{E}_{q(\mathbf{a})}[\text{log }  q(\mathbf{a})] - \mathbb{E}_{q(\mathbf{a})}[\text{log }  p(\mathbf{a} \;|\; \mathbf{b})] \\
		& = \mathbb{E}_{q(\mathbf{a})}[\text{log }  q(\mathbf{a})] - \mathbb{E}_{q(\mathbf{a})}[\text{log }  p(\mathbf{a},\mathbf{b})] \\
		& + \text{log }p(\mathbf{b})
		\label{eq:KLdivergence2}
		\end{split}
	\end{equation}

	\noindent where, for this work, $ \mathbf{a} \overset{\Delta}{=} \{\{\mathbf{f}^{(k)}\},\mathbf{Z}\} $ and $ \mathbf{b} \overset{\Delta}{=} \{\mathbf{y}\} $. However, $ p(\mathbf{b}) $ is intractable \cite{VIreview,Bull_3}; therefore, rather than the KL-divergence, an alternative object is defined, called the \emph{evidence lower bound} or \emph{elbo}. The elbo, $ \mathcal{L}_b $, is equivalent to the negative KL-divergence in Eq.\ (\ref{eq:KLdivergence2}), up to the term log $ p(\mathbf{b}) $, which is a constant with respect to $ q(\mathbf{z})$ \cite{Bull_3}, 
	
	\begin{equation}
		\begin{split}
			\mathcal{L}_{b}(\mathbf{a}) 
			& = \mathbb{E}_{q(\mathbf{a})}[\text{log }  p(\mathbf{a},\mathbf{b})] - \mathbb{E}_{q(\mathbf{a})}[\text{log }  q(\mathbf{a})] \\
			& = \int q(\mathbf{a}) \text{log} \frac{p(\mathbf{a},\mathbf{b})}{q(\mathbf{a})}\text{d}\mathbf{a}
			\label{eq:Lb1}
		\end{split}
	\end{equation}

	Maximising $ \mathcal{L}_b $ minimises the KL-divergence between $ q(\mathbf{a}) $ and $ p(\mathbf{a} \;|\; \mathbf{b}) $. Rearranging Eq.\ (\ref{eq:KLdivergence2}) and substituting in Eq.\ (\ref{eq:Lb1}) yields,

	\begin{equation}
		\text{log }p(\mathbf{b}) = \text{KL}(q(\mathbf{a}) \;||\; p(\mathbf{a} \;|\; \mathbf{b})) + \mathcal{L}_b
	\label{eq:Lb2}
	\end{equation}

	\noindent which shows that because $ \text{KL}( \cdot ) \geq 0 $ \cite{mackay2003information}, the evidence is lower-bounded by the elbo $ \mathcal{L}_b $, i.e., log $ p(\mathbf{b}) \geq \mathcal{L}_b $ \cite{Bull_3}. Substituting $ \mathbf{a} \overset{\Delta}{=} \{\{\mathbf{f}^{(k)}\},\mathbf{Z}\} $ and $ \mathbf{b} \overset{\Delta}{=} \{\mathbf{y}\} $ into Eq.\ (\ref{eq:Lb1}) yields,

	\begin{equation}
		\begin{split} 
			& \text{ log }p(\mathbf{y} \;|\; \mathbf{x}) =  \text{ log} \iint p(\{\mathbf{f}^{(k)}\},\mathbf{Z},\mathbf{y},\mathbf{x}) p(\{\mathbf{f}^{(k)}\}),p(\mathbf{Z})\text{d}\{\mathbf{f}^{(k)}\}\text{d}\mathbf{Z} \\
			& \geq \mathcal{L}_b = \iint q(\{\mathbf{f}^{(k)}\},\mathbf{Z}) \text{ log } \frac{p(\{\mathbf{f}^{(k)}\},\mathbf{Z},\mathbf{y},\mathbf{x})}{q(\{\mathbf{f}^{(k)}\},\mathbf{Z})}\text{d}\{\mathbf{f}^{(k)}\}\text{d}\mathbf{Z} \\
			& = \iint q(\{\mathbf{f}^{(k)}\},\mathbf{Z}) \text{ log } \frac{p(\mathbf{y} \;|\; \{\mathbf{f}^{(k)}\},\mathbf{Z},\mathbf{x})p(\mathbf{Z})\prod_{k=1}^{K}p(\mathbf{f}^{(k)} \;|\; \mathbf{x})}{q(\{\mathbf{f}^{(k)}\},\mathbf{Z})}\text{d}\{\mathbf{f}^{(k)}\}\text{d}\mathbf{Z} \\
			\label{eq:Lb3}
		\end{split}
	\end{equation}

	An improved bound, $ \mathcal{L}_{bc} $, that lower-bounds the marginal likelihood while also being an upper-bound on the standard variational bound $ \mathcal{L}_b $ \cite{king2006fast,Bull_3}, was implemented in this work. The improved lower bound can be expressed as \cite{lazaro2011variational,Bull_3},
	
	\begin{equation}
		\begin{split}
			& \text{ log } p(\mathbf{y} \;|\; \mathbf{x}) \geq \boldsymbol{\mathcal{L}}_{bc} \\
			& = \sum_{k=1}^{K} \left(\frac{1}{2} \;\Big|\!\Big|\; \mathbf{R}^{(k)\textsf{T}} \backslash \left(\mathbf{B}^{(k)\frac{1}{2}}\left(\mathbf{y}-\mathbf{m}^{(k)}\right)\right) \;\Big|\!\Big|\,\mystrut^2 - \sum_{i=1}^{N}\text{log }\mathbf{R}^{(k)}[i,i] \right) \;...\; \\
			& - \text{KL} \left(q(\mathbf{Z}) \;\Big|\!\Big|\; p(\mathbf{Z})\right) - \frac{1}{2} \sum_{i=1,k=1}^{N,K} \boldsymbol{\hat{\varPi}}[i,k] \text{ log} \left(2\pi\sigma^2\right) \\
			& \mathbf{R}^{(k)} \overset{\Delta}{=} \text{chol} \left(\mathbf{I} + \mathbf{B}^{(k)\frac{1}{2}} \mathbf{K}_{\mathbf{xx}}^{(k)} \mathbf{B}^{(k)\frac{1}{2}} \right) \\
			& \text{KL} \left(q(\mathbf{Z}) \;\Big|\!\Big|\; p(\mathbf{Z})\right) \overset{\Delta}{=} \sum_{i=1,k=1}^{N,K} \boldsymbol{\hat{\varPi}}[i,k] \text{ log } \frac{\boldsymbol{\varPi}[i,k]}{\boldsymbol{\hat{\varPi}}[i,k]}
			\label{eq:Lb4}
		\end{split}
	\end{equation}

	\noindent where the backslash operator solves the system of linear equations $ \mathbf{Ax} = \mathbf{B} $ for $ \mathbf{x} $ via $ \mathbf{A}\backslash\mathbf{B} $ and $ \text{chol}( \cdot ) $ is the Cholesky decomposition. The quantity  $ \mathbf{B}^{(k)} $ is an $ N \times N $ diagonal matrix with elements,

	\begin{equation}
		\mathbf{B}^{(k)} = 	\text{diag}\left(\left\{\frac{\boldsymbol{\hat{\varPi}}[1,k]}{\sigma^2}\;,\;...\;,\; \frac{\boldsymbol{\hat{\varPi}}[N,k]}{\sigma^2}\right\}\right)
	\label{eq:Estep3}
	\end{equation}

	\noindent The improved lower bound $ \mathcal{L}_{bc} $ depends only on $ p\left(\mathbf{Z}\right) $ and is independent of $ p\left(\{\mathbf{f}^{(k)}\}\right) $; as such, $ \mathcal{L}_{bc} $ can be referred to as the \emph{marginalised variational bound} \cite{LAZAROGREDILLA20121386,Bull_3}. For a given $ p\left(\mathbf{Z}\right) $, when an optimal choice for $ p\left(\{\mathbf{f}^{(k)}\}\right) $ is made, $ \mathcal{L}_{bc} $ is equivalent to $ \mathcal{L}_{b} $ \cite{LAZAROGREDILLA20121386,Bull_3}. This idea means that the bound is more stable over a variety of hyperparameters, because the improved bound is independent of $ p\left(\{\mathbf{f}^{(k)}\}\right) $ \cite{LAZAROGREDILLA20121386,king2006fast,Bull_3}. Using the improved lower bound is therefore useful in the M-step of the Expectation Maximisation (EM) scheme, when $ \mathcal{L}_{bc} $ is optimised with respect to the hyperparameters until convergence, with $ p\left(\mathbf{Z}\right) $ held fixed \cite{LAZAROGREDILLA20121386}.

	\subsubsection{E-step: Mean-field updates}
	
	A mean-field assumption was employed, and a family of variational distributions $ q(\mathbf{a}) \in \mathcal{Q} $ were defined such that each latent variable could be optimised independently, i.e., $ q(\{\mathbf{f}^{(k)}\},\mathbf{Z}) = q(\{\mathbf{f}^{(k)}\})q(\mathbf{Z}) $. This assumption allowed for the optimisation of $ \mathcal{L}_b $ (or $ \mathcal{L}_{bc} $) with respect to a given latent variable while keeping the others fixed, and iteratively updating each variable until convergence of the lower bound was reached (i.e., EM Maximisation) \cite{Bull_3}.
	
	If $ q(\{\mathbf{f}^{(k)}\}) $ and the marginals for each component $ q(\mathbf{f}^{(k)}) = \mathcal{N}(\boldsymbol{\mu}^{(k)},\boldsymbol{\Sigma}^{(k)}) $ are assumed known, $ \mathcal{L}_b $ can be analytically maximised with respect to $ q(\mathbf{Z}) $ by constraining $ q $ to be a probability density and setting the derivative of the bound to zero \cite{LAZAROGREDILLA20121386,Bull_3}, 
	
  	\begin{equation}
  		\begin{split}
		& q(\boldsymbol{Z}) = \prod_{i,k = 1}^{N,K}  	{\boldsymbol{\hat{\varPi}}[i,k]}^{\boldsymbol{Z}[i,k]} 
		\text{,        s.t.      } \boldsymbol{\hat{\varPi}}[i,k] \propto \boldsymbol{\varPi}[i,k]\text{ exp}(a_{ik})  \\
		& a_{ik} \overset{\Delta}{=} -\frac{1}{2\sigma^2}\left(\left(y_i - \boldsymbol{\mu}_i^{(k)}\right)^2 + \boldsymbol{\varSigma}^{(k)}[i,i]\right) - \frac{1}{2} \text{ log}\left(2\pi\sigma^2\right)
		\label{eq:Estep1}
		\end{split}
	\end{equation}	

	\noindent where Eq.\ (\ref{eq:Estep1}) implies that $ q(\mathbf{Z}) $ is factorised for each sample \cite{LAZAROGREDILLA20121386,Bull_3}. 
	
	Now, assuming that $ q(\mathbf{Z}) $ is known, the lower bound can be analytically maximised with respect to $ q(\{\mathbf{f}^{(k)}\}) $, as in,
	
	\begin{equation}
		\begin{split}
			& q(\mathbf{f}^{(k)}) = \mathcal{N}\left(\mathbf{f}^{(k)} \;|\; \boldsymbol{\mu}^{(k)},\boldsymbol{\varSigma}^{(k)}\right) \\
			& \boldsymbol{\varSigma}^{(k)} \overset{\Delta}{=} \left(\mathbf{K}_{xx}^{-1(k)} + \mathbf{B}^{(k)}\right)^{-1} \\
			& \boldsymbol{\mu}^{(k)} \overset{\Delta}{=} \mathbf{m}^{(k)} + \boldsymbol{\varSigma}^{(k)}\mathbf{B}^{(k)} \left(\mathbf{y} - \mathbf{m}^{(k)}\right) \\
		\end{split}
		\label{eq:Estep2}
	\end{equation}

	\noindent The best candidate $ \hat{q}(\mathbf{a}) $ that most closely approximates the true posterior distribution was found by initialising $ q(\mathbf{Z}) $ and $ q\left(\{\mathbf{f}^{(k)}\}\right) $ from their respective priors, and iteratively updating them by alternating Eqs.\ (\ref{eq:Estep1}) and (\ref{eq:Estep2}). As both updates are optimal with respect to the distribution that they compute, they are guaranteed to increase the lower bound on the log-marginal-likelihood \cite{LAZAROGREDILLA20121386,Bull_3}, computed via Eq.\ (\ref{eq:Lb3}). 
	
	\subsubsection{M-step: Hyperparameter optimisation}
	
	In the E-step, $ \mathcal{L}_{bc} $ was optimised with respect to $ p\left(\mathbf{Z}\right) $ until convergence by iterating Eqs.\ (\ref{eq:Estep1}) and (\ref{eq:Estep2}), with the hyperparameters held fixed. In the M-step, the lower bound is optimised with respect to all model hyperparameters, as in,
	
	\begin{equation}
		\Big\{\{\hat{\theta}_k\}_{k=1}^{K},\hat{\boldsymbol{\varPi}}\Big\} = \underset{\big\{\{\theta_k\}_{k=1}^{K},\boldsymbol{\varPi}\big\}}{\operatorname{argmax}} \{\mathcal{L}_{bc}\}
		\label{eq:Mstep1}
	\end{equation}
	
	\noindent while keeping the approximate posterior  $ q\left(\mathbf{Z}\right) $ fixed. The E- and M-steps are then alternated until the lower bound converges. In this work, constrained nonlinear optimisation was performed via the \emph{fmincon} function in MATLAB, and appropriate bounds were applied to the mean-function hyperparameters given prior knowledge of the modal characteristics of the helicopter blades. 
	
	\subsubsection{Predictive equations}
	
	Once learnt, the OMGP fit (or form) can be used to estimate the latent variables and functions and to inform damage detection. For training data $ \mathcal{D} $, the \emph{maximum a posteriori} (MAP) estimate can be used to categorise observations according to the most likely component $ k $, via \cite{Bull_3}, 
	
	\begin{equation}
		\hat{k}_i = \underset{k}{\operatorname{argmax}} \{\hat{\boldsymbol{\Pi}}[i,k]\}
		\label{eq:k1} 
	\end{equation}
	
	\noindent The MAP class component for new data $ \hat{k}_* $ is then defined as \cite{Bull_3}, 

	\begin{equation}
		\hat{k}_* = \underset{k_*}{\operatorname{argmax}} \{p({k}_* \;|\; \bold{x}_*,\bold{y}_*,\mathcal{D})\}
		\label{eq:k2} 
	\end{equation}
	
	\noindent where, for a set of test data, the posterior predictive class component $ k_* $ is expressed using Bayes Rule (note that to classify new data according to $ k_* $, both $ \mathbf{x_*} $ and $ \mathbf{y_*} $ must be observed \cite{Bull_3}),
	
	\begin{equation}
		p({k}_* \;|\; \bold{x}_*,\bold{y}_*,\mathcal{D}) =
		\frac{{p(\bold{y}_* \;|\; \bold{x}_*,k,\mathcal{D})p(k_*)}}
		{{p(\bold{y}_* \;|\; \bold{x}_*,\mathcal{D})}}
		\label{eq:PredictiveEqs}
	\end{equation}
	
	\noindent The numerator of Eq.\ (\ref{eq:PredictiveEqs}), or unnormalised posterior, is defined as \cite{Bull_3},
	
 	\begin{equation}
		p(\bold{y}_* \;|\; \bold{x}_*,k,\mathcal{D})p(k_*)
		\overset{\Delta}{=} 
		\mathcal{N}(\bold{y}_* \;|\;
		\boldsymbol{\mu}_*^{(k)},\boldsymbol{\varSigma}_*^{(k)})
		\boldsymbol{\varPi}[*,k]
		\label{eq:PredictiveEqs1}
	\end{equation}
	
	\noindent and the denominator of Eq.\ (\ref{eq:PredictiveEqs}), which is the marginal likelihood (evidence), is \cite{Bull_3},
	
	\begin{equation}
		\begin{split}
		p(\bold{y}_* \;|\; \bold{x}_*, \mathcal{D}) & \approx \sum_{k=1}^{K} \boldsymbol{\varPi}[*,k] \int p\left(\bold{y}_* \;|\; \mathbf{f}^{(k)},\bold{x}_*, \mathcal{D}\right)q\left(\mathbf{f}^{(k)} \;|\;\mathcal{D}\right)\text{d}\mathbf{f}^{(k)} \\
		& = \sum_{k=1}^{K}\mathcal{N}\left(\bold{y}_* \;|\; \boldsymbol{\mu}_*^{(k)},\boldsymbol{\varSigma}_*^{(k)}\right)\boldsymbol{\varPi}[*,k] \\
		\label{eq:PredictiveEqs2}
		\end{split}
	\end{equation}
	
	\noindent where \cite{Bull_3},
	
	\begin{equation}
		\begin{split}
		& \boldsymbol{\mu}_*^{(k)} \overset{\Delta}{=} \mathbf{m}_*^{(k)} + \mathbf{K}^{(k)}_\mathbf{{x_*x}}\left(\mathbf{K}^{(k)}_\mathbf{{xx}} + \mathbf{B}^{(k)-1}\right)^{-1}\left(\mathbf{y} - \mathbf{m}^{(k)}\right) \\
		& \boldsymbol{\Sigma}_*^{(k)} \overset{\Delta}{=} \mathbf{K}^{(k)}_\mathbf{{x_*x_*}} - \mathbf{K}^{(k)}_\mathbf{{x_*x}}\left(\mathbf{K}^{(k)}_\mathbf{{xx}} + \mathbf{B}^{(k)-1}\right)^{-1} \mathbf{K}^{(k)}_\mathbf{{xx_*}} + \mathbf{R}_*^{(k)} \\
		& \mathbf{R}_*^{(k)} \overset{\Delta}{=} \sigma^2 \mathbf{I}_\mathbf{M}\\
		\label{eq:PredictiveEqs3}
		\end{split}
	\end{equation}
	
	The prior mixing proportions for new observations, $ \boldsymbol{\varPi}[*,k] $, weight each component equally \emph{a priori}, such that the sum of the weights is equal to 1, or $ \boldsymbol{\varPi}[*,k] = 1/K $ \cite{Bull_3}. The predictive equations for the OMGP are quite similar to those for the conventional GP in Eq.\ (\ref{eq:GPposterior}), with the exception of the noise component for the training data $ \mathbf{B}^{(k)-1} $ which is scaled for each sample according to the inverse of the mixing proportion  $ \boldsymbol{\hat{\varPi}}[i,k]^{-1} $ \cite{LAZAROGREDILLA20121386,Bull_3}. As such, the contribution of each sample is weighted with respect to its posterior predictive component in the mixture \cite{Bull_3}. 
	
	\subsubsection{Practical implementation of the described technology}
	
	In this work, a population form was developed for the helicopter blades using a mixture of Gaussian processes, and new (simulated) FRF data were used to evaluate the sensitivity of the form for novelty detection. There were four healthy helicopter blades and therefore four classes ($ K = 4 $). Because the new data had an equal chance of belonging to any of the four classes, the prior mixing proportion weighted each class equally, with $ \boldsymbol{\varPi}[*,k] = 0.25 $. The marginal likelihood (evidence), defined in Eq.\ (\ref{eq:PredictiveEqs2}) was used as a novelty index, and provided an averaged likelihood of new data belonging to any of the $ k $ classes. For computational ease and to avoid numerical underflow, the calculations were performed in log space. Specifically, negative log quantities were used, such that the visual presentation of data novelty was consistent with the Mahalanobis squared-distance (MSD) novelty index from \cite{Bull_1,Bull_2,WORDEN2000647}. As such, data below a certain threshold were considered normal or \emph{inyling}, and data above the threshold were considered novel or \emph{outlying}.
	
	A suitable threshold for novelty detection was determined using a similar method as with the MSD in \cite{Bull_1,Bull_2,WORDEN2000647}. Bootstrap-sampling was used, where 1000 samples were randomly selected from the normal-condition data used to train the form. The negative log-marginal-likelihood was then calculated for the samples, for a large number of trials. The critical value was the threshold with a certain percentage of the calculated values below it. This work used a 99\% confidence interval, or $ 2.58\sigma $. 
	
	\subsubsection{Model evaluation}
	
	The normalised mean squared-error (NMSE) and Mahalanobis squared\--dist\-ance (MSD) were calculated to compare the different models developed in this work (e.g., supervised versus unsupervised mixture of GPs), particularly when it was not feasible to use the model's objective function as a performance metric (e.g., when comparing two models learnt using different training data). The NMSE was computed to assess each test observation against the mean of its most likely class component $ k_* $, via \cite{Bull_3}, 
	
	\begin{equation}
		\text{NMSE} = \frac{100}{M\sigma^2_{\mathbf{y_*}}}\left(\boldsymbol{\mu}^{(\hat{k}_*)}_* - \mathbf{y_*}\right)^\textsf{T}\left(\boldsymbol{\mu}^{(\hat{k}_*)}_* - \mathbf{y_*}\right)
		\label{eq:NMSE}
	\end{equation}

	Likewise, the (normalised) MSD was computed via,
	
	\begin{equation}
		\text{MSD} = \frac{1}{M}\sum_{i=1}^{M}\frac{\left(\boldsymbol{\mu}^{(\hat{k}_*)}_*[i] - \mathbf{y_*}[i]\right)^2}{\boldsymbol{\varSigma}^{(\hat{k}_*)}_*[i,i]}
		\label{eq:MSD}
	\end{equation}

	\noindent where $ M $ is the number of test data points (not included in training set). In this work, the MSD was used only to assess the predictive variance $ \boldsymbol{\varSigma}^{(k)}_* $. The MSD is less useful when assessing the fit of the OMGP, as the error is scaled by the predictive variance $ \boldsymbol{\varSigma}^{(k)}_* $ \cite{Bull_3}.

\vspace{12pt} 
\section{Experimental campaign on a population of helicopter blades}
\label{}
	Vibration data were collected over a series of tests from a set of four healthy, full-scale composite helicopter blades, in fixed-free and free-free configurations. This section discusses how differences among the blades and changes in boundary conditions present as changes in the FRFs of the blades, particularly with respect to the locations of the peaks. 
	
	\subsection{Experiments}
	Vibration data were collected at ambient temperature from four healthy, full-scale composite helicopter blades using Siemens PLM LMS SCADAS hardware and software at the Laboratory for Verification and Validation (LVV) in Sheffield, UK. Testing was performed in fixed-free and free-free configurations. To approximate a fixed-free boundary condition, the root end of each blade was placed in a substantiated strong-wall mount. To approximate a free-free boundary condition, the blade was suspended from a gantry via springs and cables. 
	
\begin{figure}[!h]
		\centering
		\subfloat[\label{fig:blade_fixed}]{\includegraphics[width=\textwidth, trim = {0 6cm 2cm 8cm},clip]{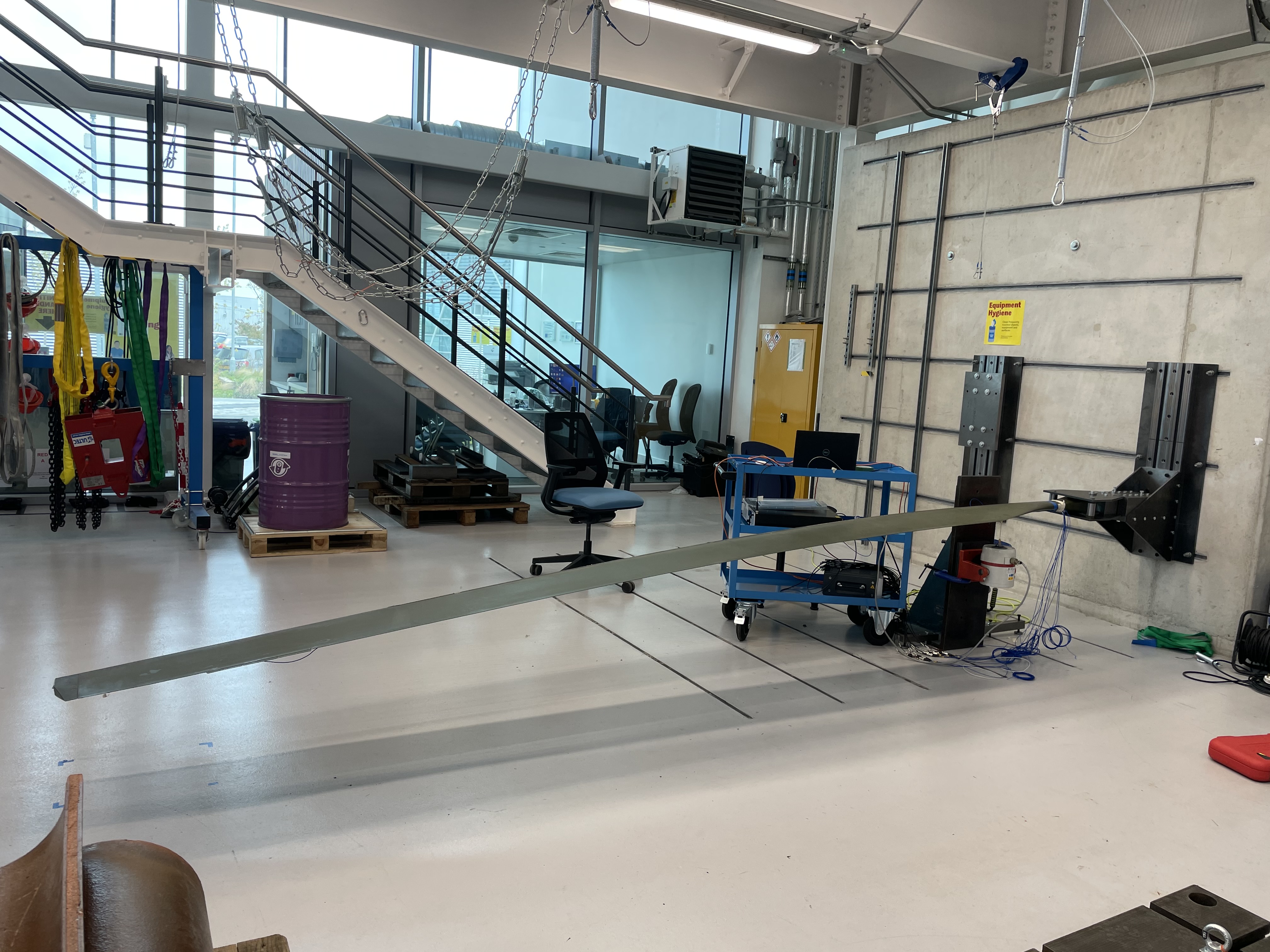}} \\
		\subfloat[\label{fig:blade_free}]{\includegraphics[width=\textwidth, trim = {1cm 6cm 1cm 8cm},clip]{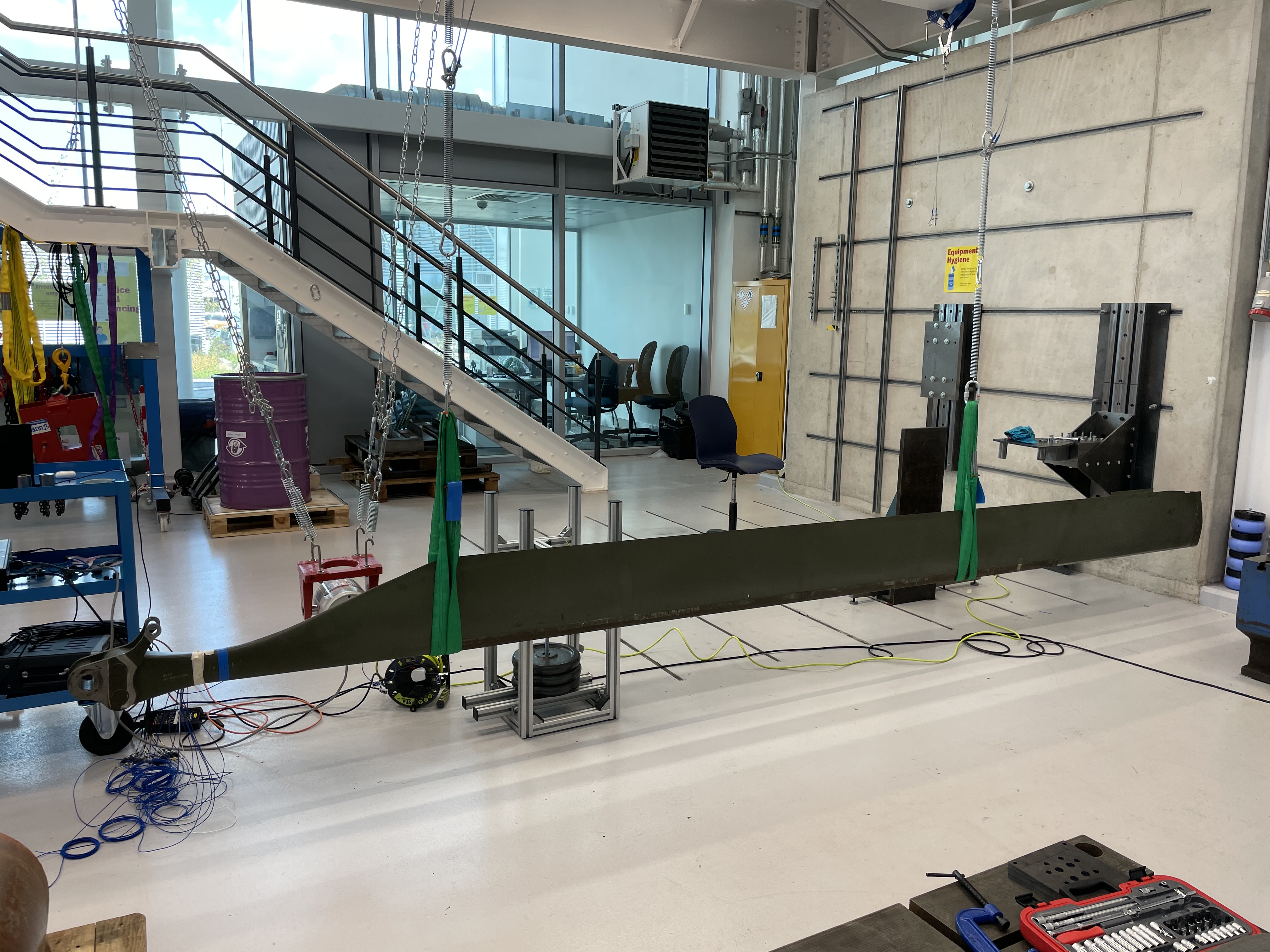}}
		\caption{Helicopter blade in (a) fixed-free and (b) free-free configurations.}
	\end{figure}

\begin{figure}[!h]
		\centering
		{\includegraphics[width=\textwidth, trim = {2.8cm 0cm 2.5cm 0cm},clip]{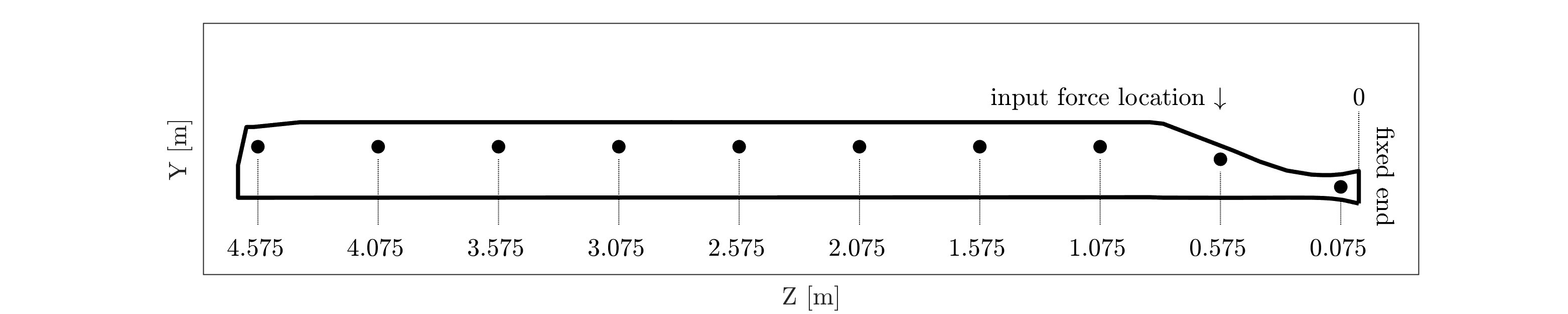}}
		\caption{Sensor locations on helicopter blades.}
		\label{fig:blade_sensors}
	\end{figure}

	Ten uniaxial 100 mV/g accelerometers were placed along the length of the underside of each blade. Note that the same accelerometers were used on each blade, and care was taken to ensure that they were attached to approximately the same locations on each blade. For the fixed-free tests, an electrodynamic shaker with force gauge was mounted to a fixture bolted to the laboratory floor and attached to the blade 0.575 metres from the fixed end. The shaker was attached to the underside of the blade in the flapwise direction. A continuous random excitation was generated in LMS (note: LMS refers to Siemens PLM LMS SCADAS hardware and software) and applied to excite the blade up to 400 Hz. Approximately 7.4 minutes of throughput force and acceleration data were collected for each test, with a time step of 1.25e-03 seconds and a sample rate of 800 Hz. The data were then divided into 20 blocks, each with a dimension of 16384. Some data at the start and end of the acquisition were discarded as these data corresponded to powering up and down of the shaker. A Hanning window was applied to each data block and FRFs were computed in LMS. The FRFs were then averaged in the frequency domain. For the free-free tests, the shaker was connected at the same location on the blade as the fixed-free tests, but was suspended from a gantry via springs. A continuous random excitation was generated in LMS and applied to excite the blade up to 512 Hz. Approximately 5.8 minutes of throughput force and acceleration data were collected for each test, with a time step of 9.77e-04 seconds and a sample rate of 1024 Hz. As with the fixed-free tests, the data were divided into 20 blocks, each with a dimension of 16384, and some data corresponding to powering up/powering down of the shaker were discarded. A Hanning window was applied to each data block and FRFs were computed for each measurement and then averaged in the frequency domain. The measured bandwidths for the fixed-free and free-free tests were selected to maximise the number and accuracy of the captured modes (there was a large dip in the input spectrum just after 400 Hz for the fixed-free tests and just after 512 Hz for the free-free tests). Data were collected over a relatively-long time period for each test and a large number of averages were used to improve the spectra at the lower modes near the limits of the operating range of the sensors. The experimental setup for the fixed-free and free-free tests are shown in Figures \ref{fig:blade_fixed} and \ref{fig:blade_free}, respectively. The sensor layout is shown in Figure \ref{fig:blade_sensors}. Acquisition parameters are listed in Table \ref{tab:blade_acquisition}. 
	
\begin{figure}[!h]
		\centering
		\subfloat[\label{fig:representative_spectra_fixed}]{\includegraphics[width=1\textwidth]{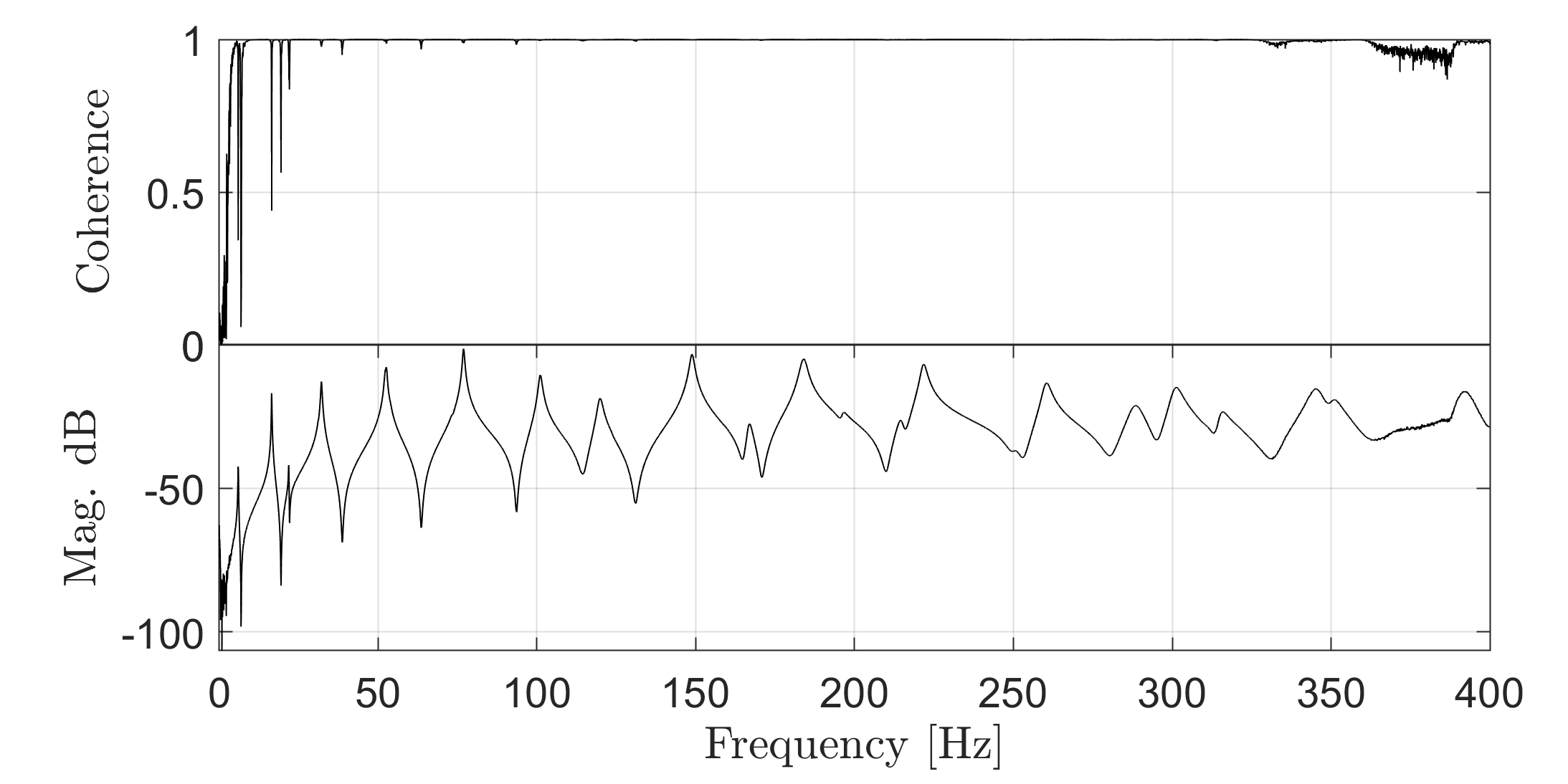}} \\
		\subfloat[\label{fig:representative_spectra_free}]{\includegraphics[width=1\textwidth]{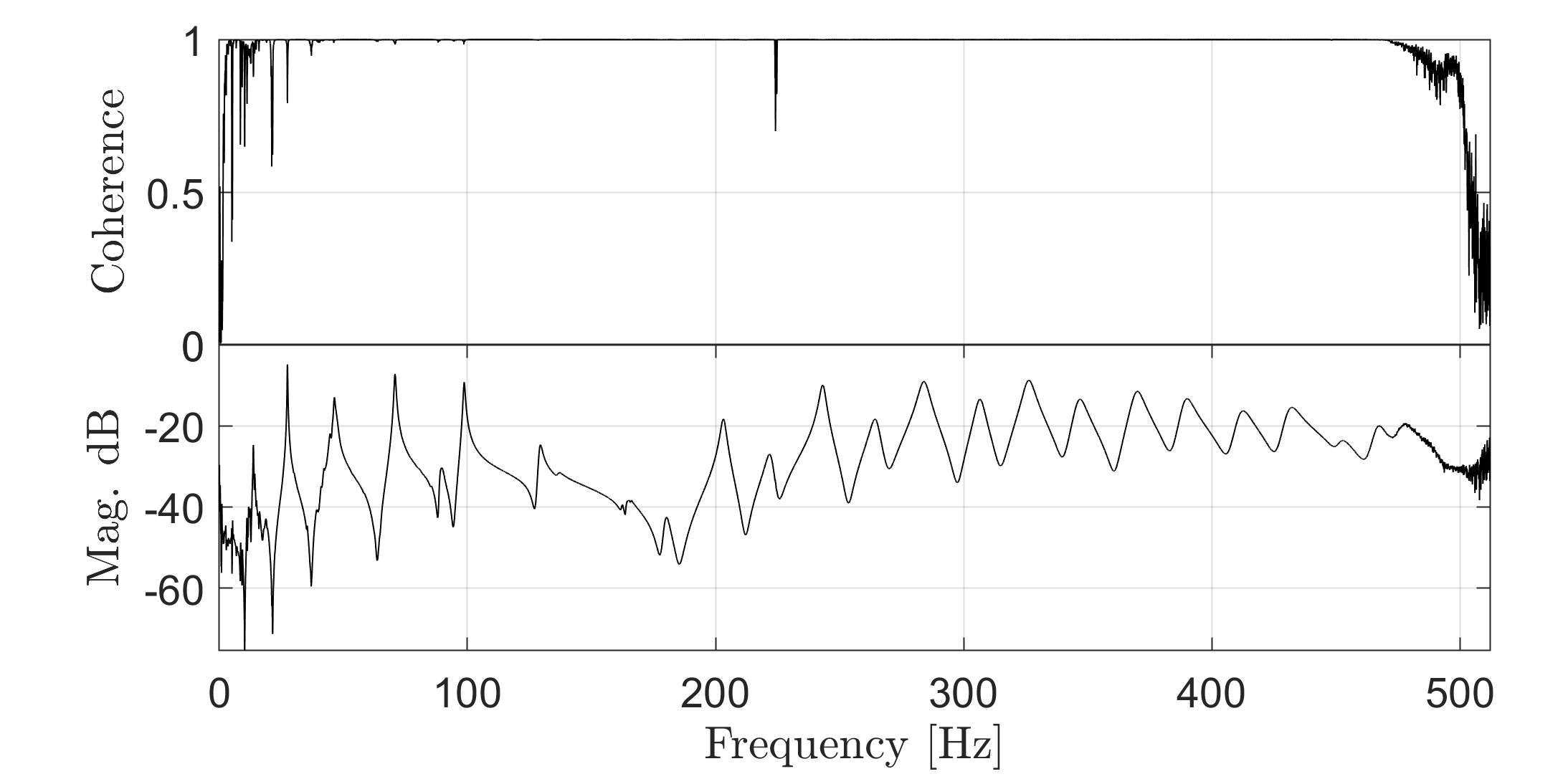}}
		\caption{Representative drive-point coherence and FRF from Blade 3 in (a) fixed-free and (b) free-free configurations.}
		\vspace{18pt}
	\end{figure}

	An example drive-point coherence spectrum and FRF for Blade 3 in fixed-free and free-free conditions (all computed from data obtained at the accelerometer located near the force-input location, 0.575 metres from the blade root) are shown in Figures \ref{fig:representative_spectra_fixed} and \ref{fig:representative_spectra_free}, respectively. These spectra are representative of all measurements and show that coherence values were close to 1 at the resonances, with the exception of some difficulty with collecting the first two modes below 10 Hz. This difficulty in acquiring the first few modes resulted in part from the operating frequency range of the accelerometers and the extremely high flexibility of the blades, which limited the shaker placement.
	
	\begin{table}[!h]
		\centering
		\caption{Acquisition parameters for vibration testing of helicopter blades.\label{tab:blade_acquisition}}
	\begin{tabular}{ p{4cm} p{2.25cm} p{2.25cm} }
		\hline
		Acquisition Settings & Fixed-Free & Free-Free \\
		\hline
		Time step   & 1.25e-03 s & 9.77e-04 s  \\
		Acquisition time     & 20.48 s & 16.00 s \\
		Bandwidth  & 400 Hz & 512 Hz \\
		Lines & 8192  & 8192 \\
		Frequency step  &  4.88e-02 Hz &  6.25e-02 Hz \\
		Window & Hanning & Hanning \\
		\hline
	\end{tabular}
	\end{table}

	\subsection{Discussion of experimental results}
	\label{}

\begin{figure}[h!]
	\centering
	\subfloat[\label{fig:FRFs_allblades_fixed}]{\includegraphics[width=1\textwidth]{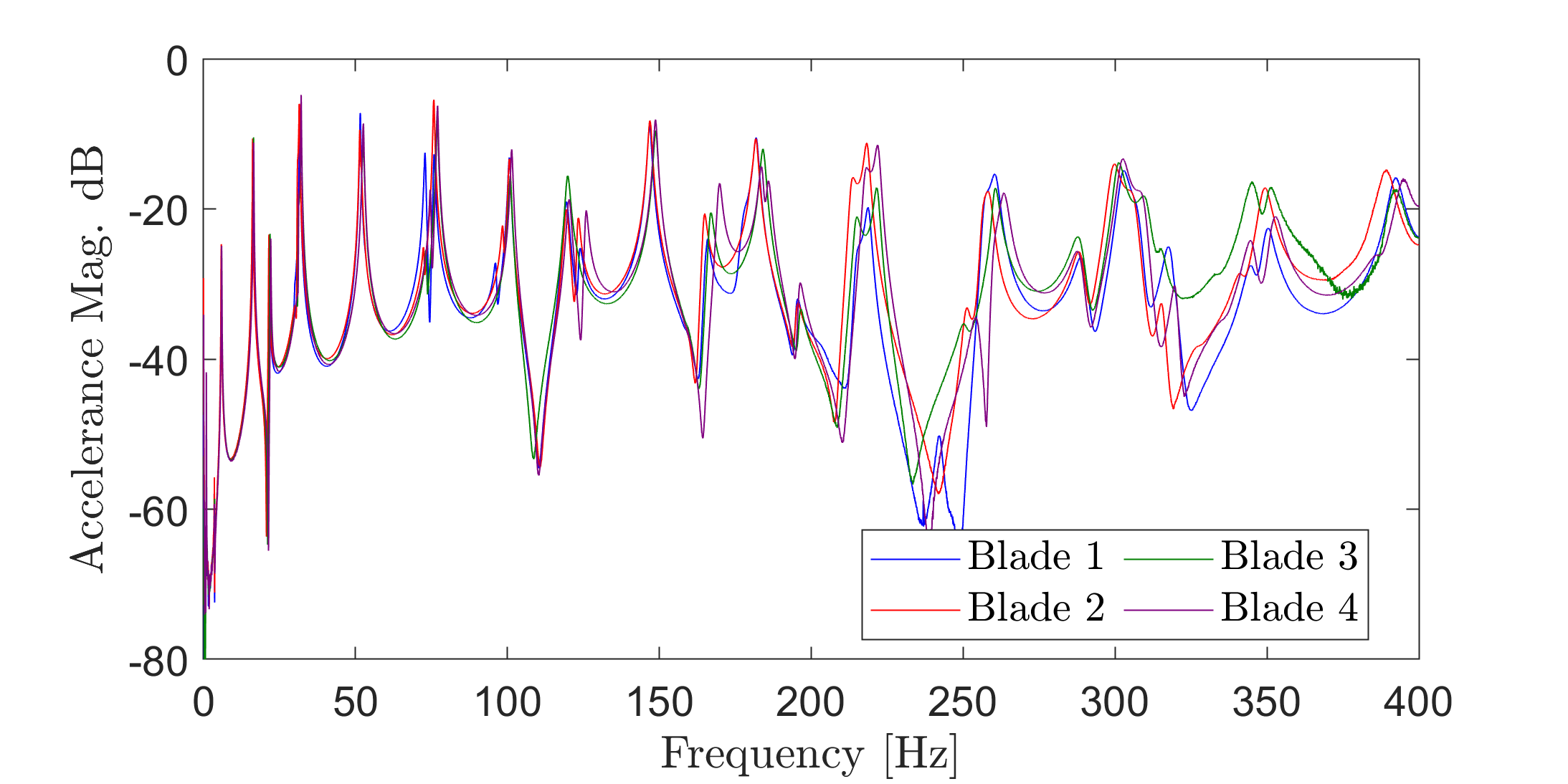}} \\
	\subfloat[\label{fig:FRFs_allblades_fixed_zoomed}]{\includegraphics[width=1\textwidth]{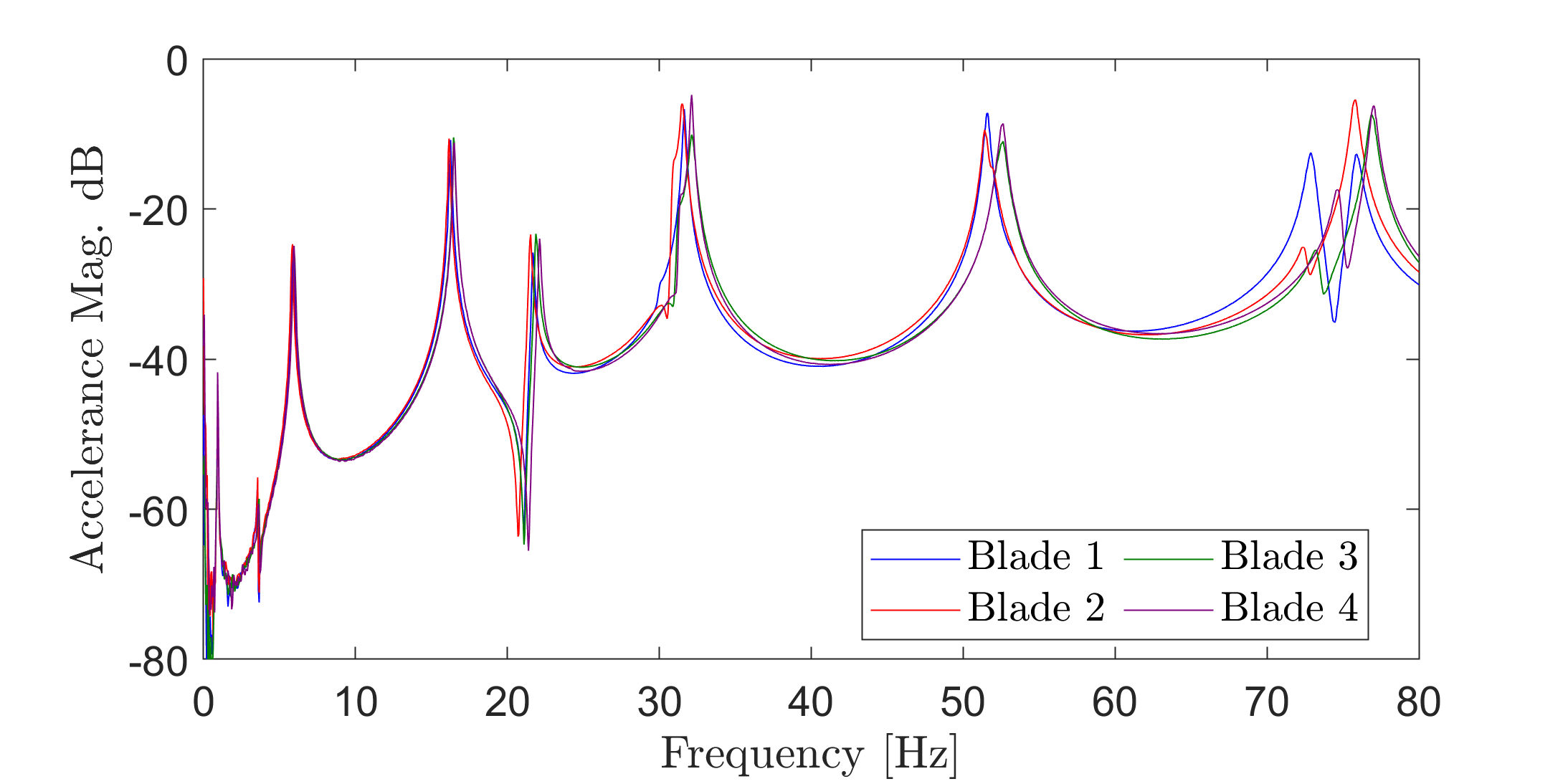}} \\ 
	\caption{FRF magnitudes for the helicopter blades in fixed-free (a) full bandwidth and (b) first 80 Hz.}
	\end{figure}

\begin{figure}[h!]
	\centering	\subfloat[\label{fig:FRFs_free}]{\includegraphics[width=1\textwidth]{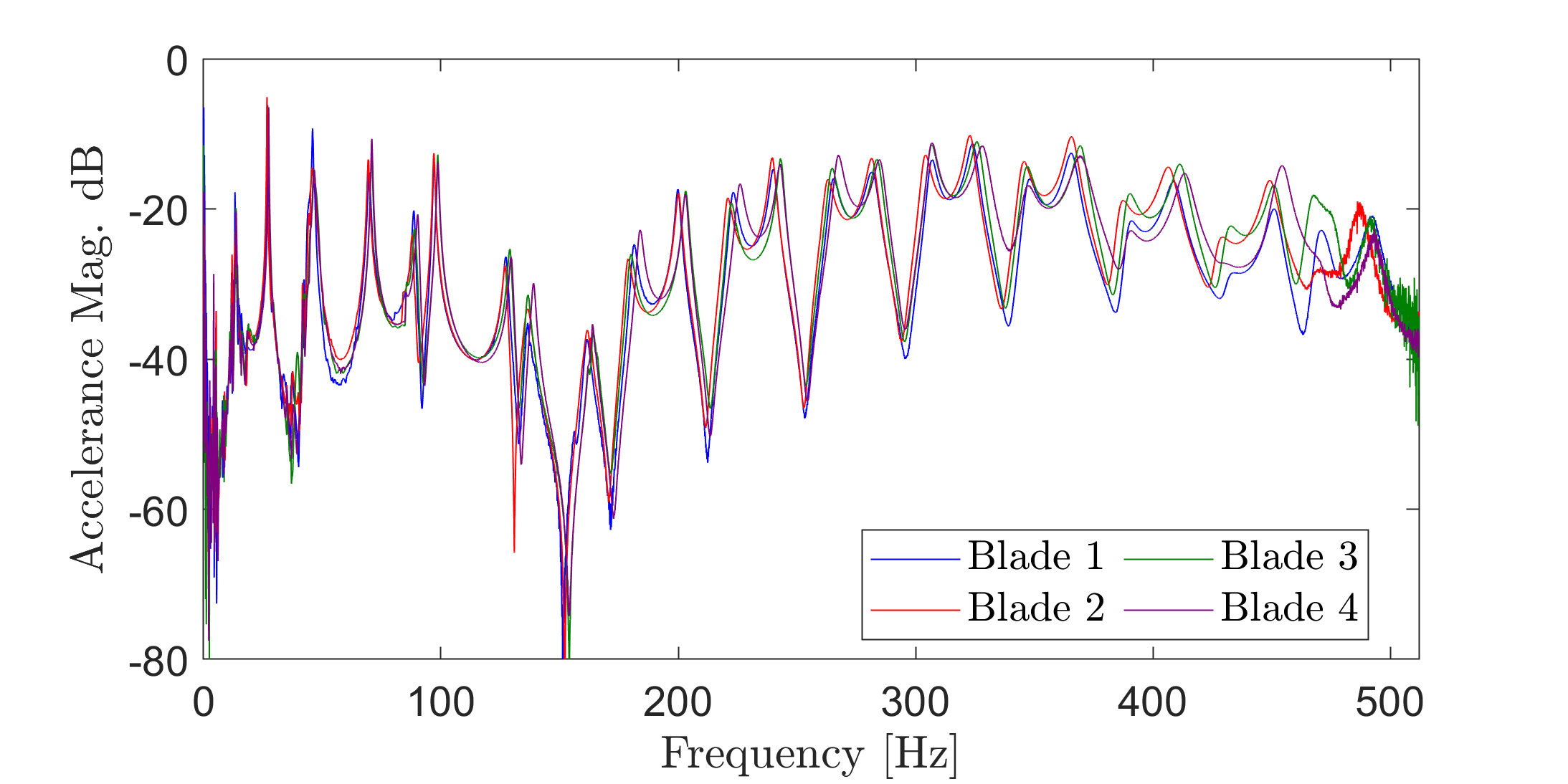}} \\	\subfloat[\label{fig:FRFs_free_zoomed}]{\includegraphics[width=1\textwidth]{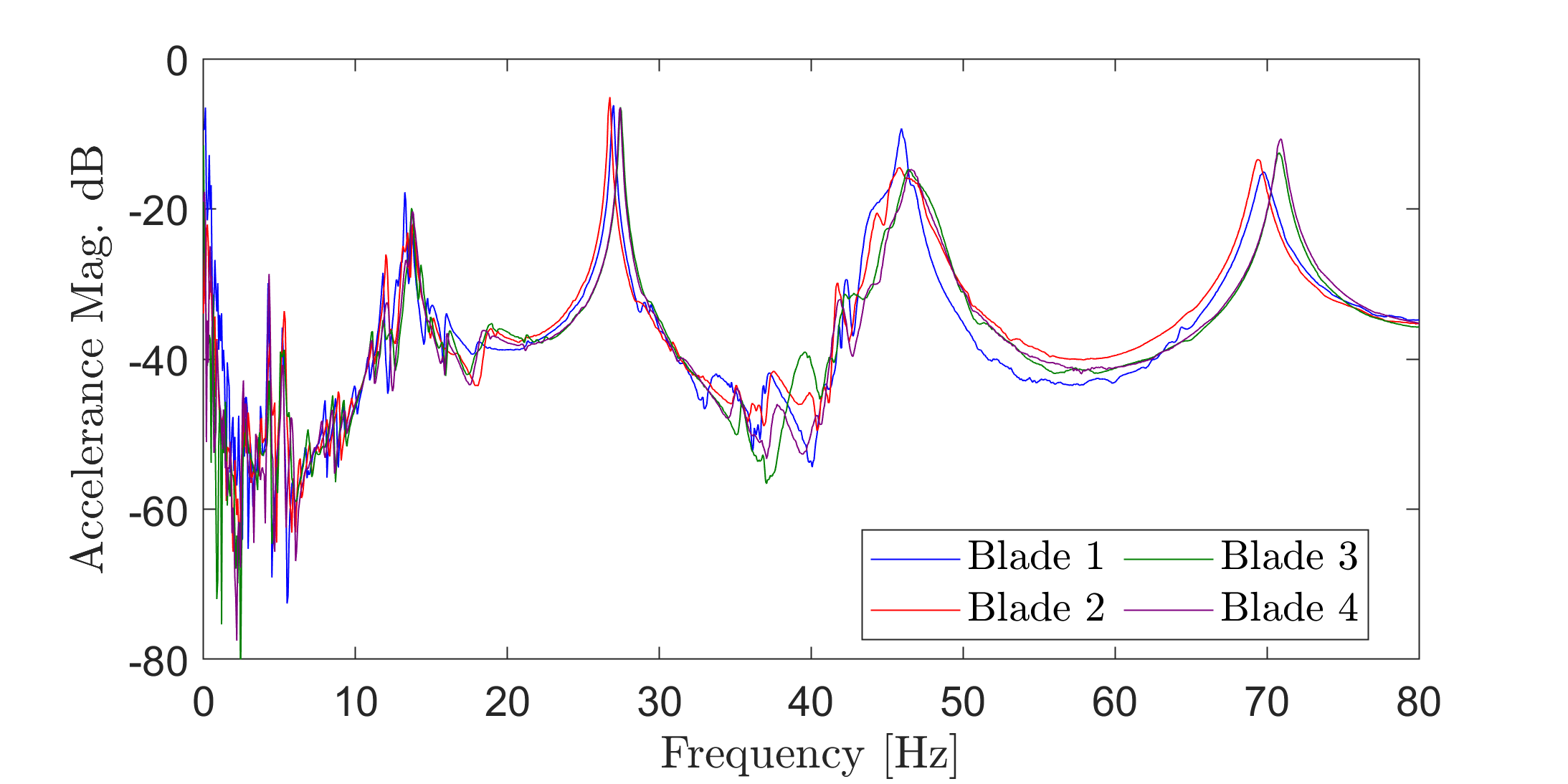}} 
	\caption{FRF magnitudes for the helicopter blades in free-free (a) full bandwidth and (b) first 80 Hz.}
	\end{figure}

	This work is concerned with demonstrating how differences among the nom\-inally-identical blades (with consideration for boundary condition effects), pre\-sent as changes in the peak positions as visible in the FRFs. Data were collected from 10 accelerometers during each test. This discussion considers the results from the accelerometer near the blade tip, as the feature changes visible in these spectra are representative of all test points along the blade length. The averaged FRFs for each blade in the fixed-free condition are shown in Figures \ref{fig:FRFs_allblades_fixed} and \ref{fig:FRFs_allblades_fixed_zoomed}. Figure \ref{fig:FRFs_allblades_fixed} shows the full measured 400 Hz bandwidth and Figure \ref{fig:FRFs_allblades_fixed_zoomed} shows only modes below 80 Hz. Likewise, the averaged FRF for each blade in free-free is shown in Figures \ref{fig:FRFs_free} and \ref{fig:FRFs_free_zoomed}, where Figure \ref{fig:FRFs_free} shows the full measured 512 Hz bandwidth and Figure \ref{fig:FRFs_free_zoomed} shows only modes below 80 Hz.

	Figures \ref{fig:FRFs_allblades_fixed} and \ref{fig:FRFs_allblades_fixed_zoomed} show increasing variability with respect to frequency, which is an expected result, given that higher-frequency modes are more sensitive to small physical changes than lower-frequency modes. For modes less than 80 Hz, the maximum frequency difference among the blades was approximately 2.5 Hz, while for modes greater than 80 Hz, the maximum frequency difference was approximately 6.3 Hz. Figures \ref{fig:FRFs_free} and \ref{fig:FRFs_free_zoomed} show a very different distribution of peaks compared to the fixed-free tests (as expected, given such a dramatic change in boundary condition), but also 
	show similar variation among the blades. For modes less than 80 Hz, the maximum frequency difference among the blades was approximately 1.6 Hz. For modes greater than 80 Hz, the maximum frequency difference among the blades was approximately 6.8 Hz. 
	
	Note that both the fixed-free and free-free results show grouping at several of the peaks, where Blades 1 and 2 appear closely aligned in frequency while Blades 3 and 4 appear closely aligned. These results are particularly noticeable in Figure \ref{fig:FRFs_allblades_fixed_zoomed}, between 40 Hz and 60 Hz, and in Figure \ref{fig:FRFs_free_zoomed}, between 60 Hz and 80 Hz. These findings are quite relevant for PBSHM, which seeks to transfer valuable information across similar structures \cite{gardner2021foundations}. If these data were used to develop a normal condition for the blades against which to evaluate new data for novelty, this variation could present some interesting problems (e.g., negative transfer). For example, if Blade 4 experienced some damage that reduced its natural frequency, such that it aligned with the same peak from Blade 1, the damage may be missed. Another potential issue would be if a new blade were added to the population and was slightly less stiff than Blade 2, the new blade may be flagged as damaged when compared to the normal condition developed using the original blades and test results. The development of a population form to represent the behaviour of the group is discussed in the next section. Only the fixed-free results were used in developing the form, although significant changes in boundary conditions could be addressed using techniques such as transfer learning \cite{gardner2021foundations}. Both the fixed-free and free-free tests showed considerable variability among the blades and a similar grouping pattern.

\vspace{12pt} 
\section{Development of a Gaussian process form}
\label{}
	To differentiate between normal and damaged states, an important development in PBSHM will involve establishing a normal condition for a population of structures. In certain cases, the behaviour of the group can be represented using a general model, called a \emph{form}, against which new data can be evaluated for novelty \cite{Bull_1,Bull_2,Bull_3}. This work used Gaussian process (GP) regression to model the form for the FRFs of the helicopter blades in fixed-free (recall that the free-free tests showed similar variability among the blades as the fixed-free tests). 

	Two examples are discussed. In the first example, the form was developed using a supervised mixture of GPs with normal-condition experimental data from the blades. This example was supervised because it was assumed \emph{a priori} which data belonged to each helicopter blade. In the second example, the normal-condition form was developed using an (unsupervised) OMGP, where the number of trajectory functions were assumed known \emph{a priori}, but it was not assumed which data belonged to a given helicopter blade, representing a situation with incomplete label information. This situation often occurs in SHM, when the labels correspond to a range of operational, environmental, or damage conditions for each structure. 
	
	\vspace{12pt} 
	\subsection{A form using a supervised mixture of Gaussian processes}
	A supervised mixture of Gaussian processes assumes that the data labels are known \emph{a priori}. In this case, it was assumed to be known which FRF data were associated with each helicopter blade. This assumption allowed for each GP in the mixture to be treated independently using the equations for a conventional GP (Eqs.\ (\ref{eq:latentfunc}-\ref{eq:optimisation})). While in a testing environment, FRFs are often described via magnitude and phase; however, the uncertainty distribution of the magnitude is non-Gaussian. As such, it is not proper to fit the magnitude directly using GPs (doing so will result in support for negative values, but $ \text{mag}[\text{H}_{ij}] \geq 0 $). If the time domain data are random realisations of a stochastic process, the fast Fourier transforms (FFT) of the signals are random realisations of the spectra, because the FFT is a linear operator \cite{bendat2011random}. As quotients of FFTs (rather, quotients of cross-spectra and input auto-spectra), the real and imaginary parts of the FRF are not strictly Gaussian; however, for this paper it is assumed that each can be approximated as GPs. In this work, the FRF was expressed as real and imaginary parts, and the real and imaginary parts of the FRF were fit separately via Eqs.\ (\ref{eq:latentfunc}-\ref{eq:optimisation}). Note that several conference papers related to this research \cite{DardenoIWSHM1,DardenoIWSHM2}, provided a preliminary approach where magnitude was fitted directly. Although the resulting form was useful for damage detection, the support for negative magnitudes does not make sense and was therefore not employed in later work.
	
	To develop the form, a narrow frequency band was selected between 48 and 56 Hz, containing the fifth bending mode of the blades. Closer inspection of the data revealed a likely second mode in the same frequency range; however, the modes were so closely-spaced that they could conceivably be a slightly distorted single mode. An SDOF assumption was imposed, and an SDOF mean function was applied to the GPs. In other words, the analysis was treated as a grey box, in that some physics were imposed via the mean function. Simplifying the analysis to include a single mode in the band of interest was a reasonable assumption, as most of the physics (and variation), were accounted for with the SDOF mean function, and the remaining distortion was resolved via the black box component of the GP.  (Note that the decision to focus on a narrow frequency band was made to simplify the analysis while demonstrating the proposed technology, although the form could be developed over a larger band and with a multiple DOF assumption).

\begin{figure}[h!]
		\centering
		\subfloat[\label{fig:blades_GP_real_EX1a}]{\includegraphics[width=1\textwidth]{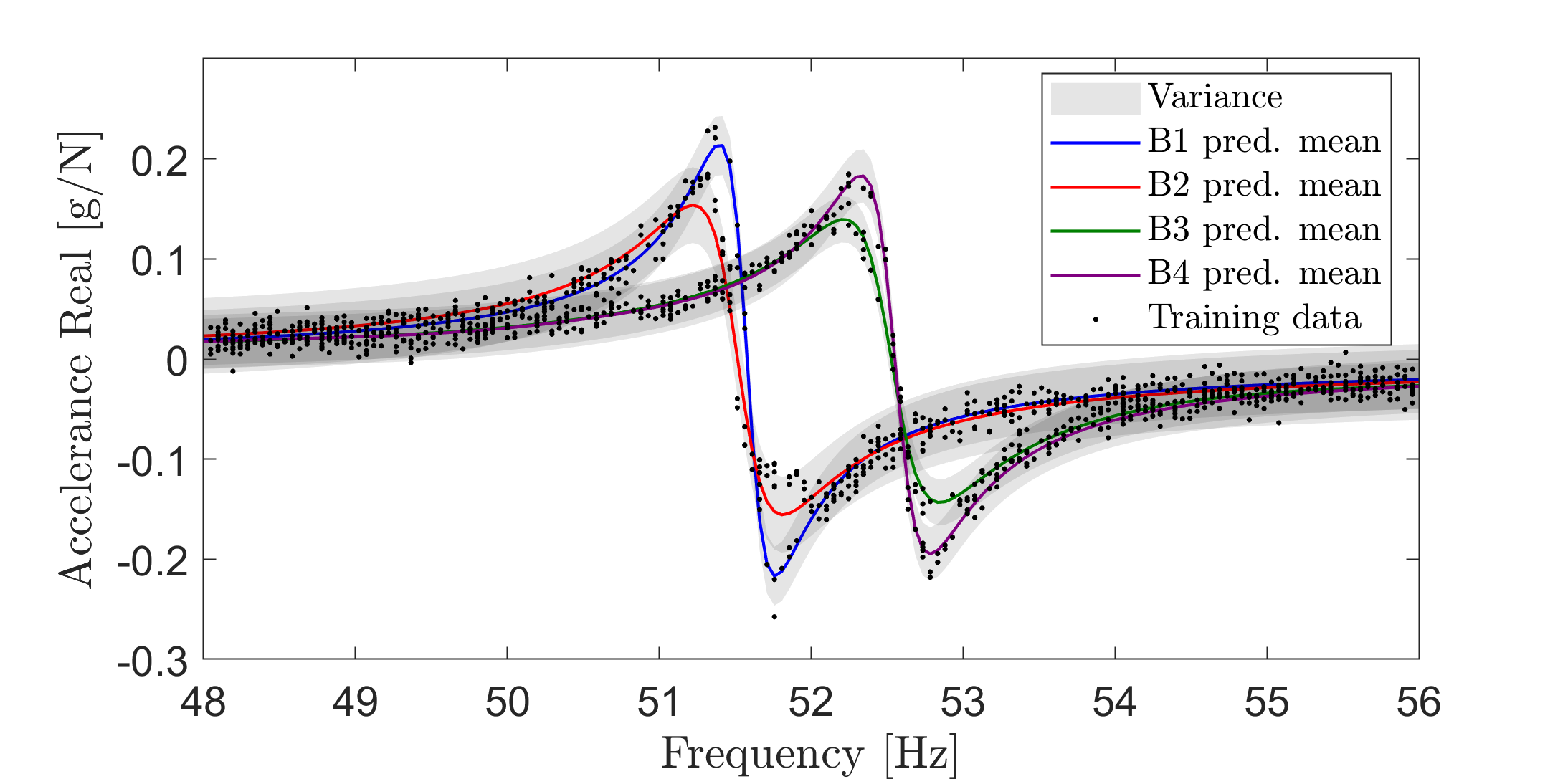}} \\
		\subfloat[\label{fig:blades_GP_imag_EX1a}]{\includegraphics[width=1\textwidth]{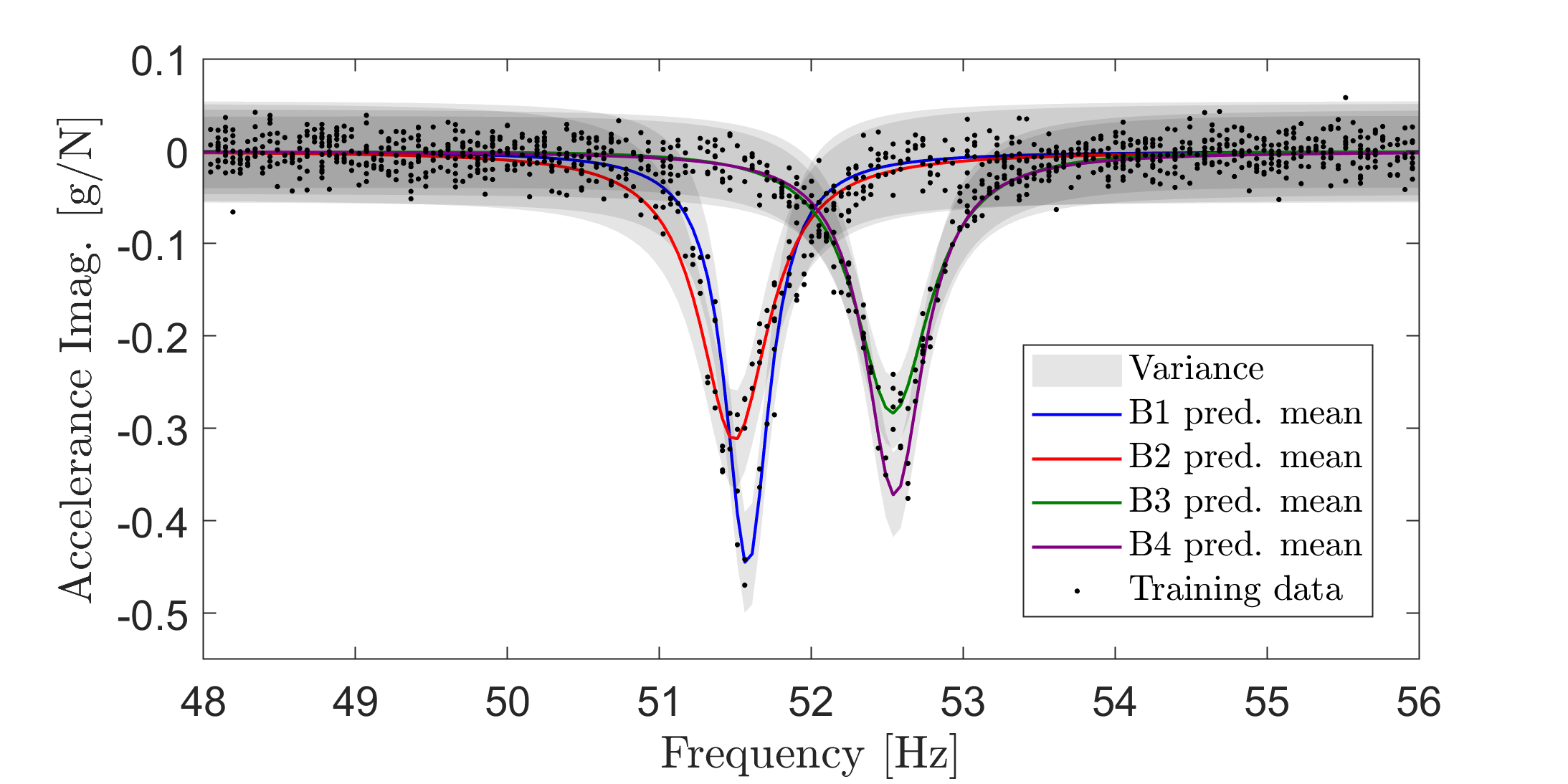}} \\ [-0.5ex]
		\caption{Predictions for supervised mixture of GPs using (a) real and (b) imaginary FRF data.}
	\end{figure}

\begin{figure}[h!]
		\centering
		\subfloat[\label{fig:blades_GP_real_EX1b}]{\includegraphics[width=1\textwidth]{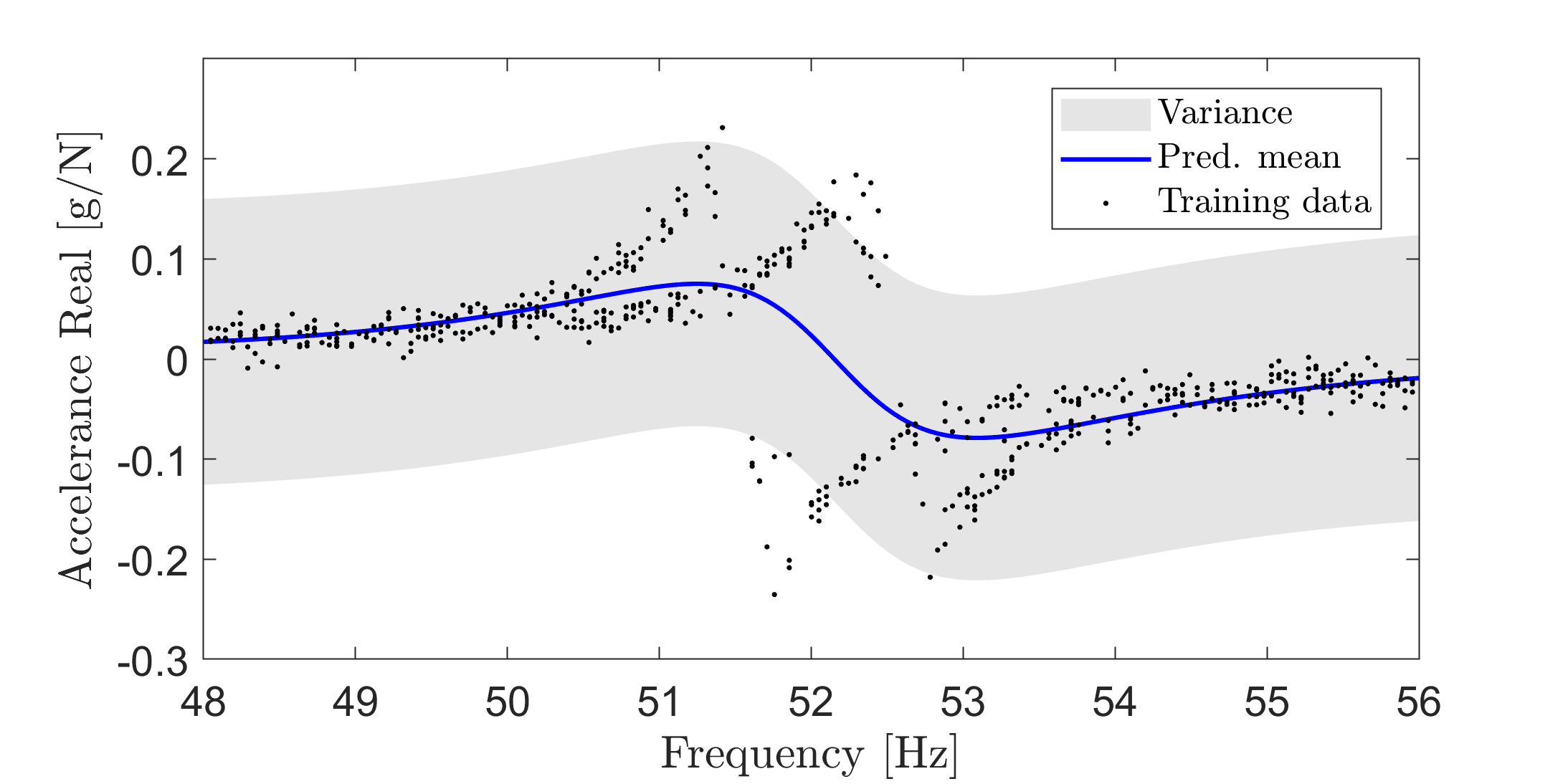}} \\
		\subfloat[\label{fig:blades_GP_imag_EX1b}]{\includegraphics[width=1\textwidth]{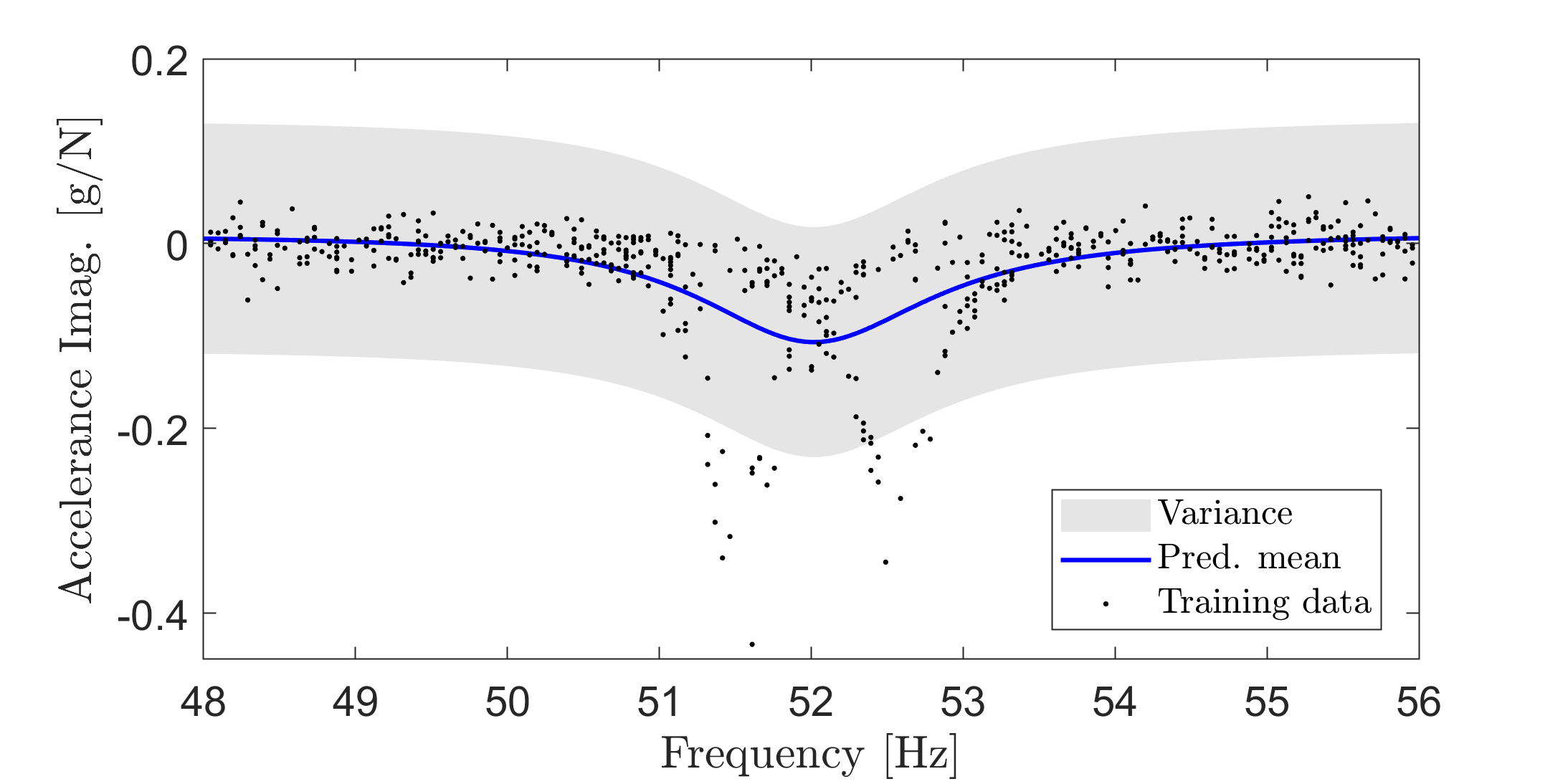}} \\ [-0.5ex]
		\caption{Predictions for single GP using (a) real and (b) imaginary FRF data.}
	\end{figure}

	To fit the real part, the real part of the FRFs from the fixed-free tests from each blade was copied 20 times, and random Gaussian noise with a magnitude equal to 5\% of the absolute peak value of each FRF was added to each copy. The data copies were then concatenated into a vector, and 300 training points were randomly selected from the vector. The modal damping FRF model from Eq.\ (\ref{eq:modalFRF}) was used to generate a mean function, and the modal damping, residue, and natural frequency were used as the mean-function hyperparameters. Upper and lower bounds were applied to constrain the hyperparameter optimisation, based on prior knowledge of the blades and visual inspection of the data (e.g., the natural frequency was bounded between 40 and 60 Hz). In addition, the sign of the data ($ + / - $) was assumed to be known \emph{a priori}, and can easily be determined via data inspection. This process was repeated using the imaginary part of the FRFs, and was repeated for each blade. The posterior predictive mean of the GP is plotted with the mean function, variance, and training data for the real and imaginary FRFs in Figures \ref{fig:blades_GP_imag_EX1a} and \ref{fig:blades_GP_real_EX1a}. For comparison, a single GP was learnt for the same data (although using a different training set comprised of 600 data points), as shown in Figures \ref{fig:blades_GP_imag_EX1b} and \ref{fig:blades_GP_real_EX1b}. 

	Figures \ref{fig:blades_GP_imag_EX1a} and \ref{fig:blades_GP_real_EX1a} show that the mixture provided a reasonable fit for the FRF data, and the majority of the training data were enclosed by the variance bounds. Furthermore, the NMSE was calculated using Eq.\ (\ref{eq:NMSE}) using an independent test data set ($ M = 12480 $), and showed acceptable values (2.4 for the real fit, and 6.9 for the imaginary fit). Conversely, Figures \ref{fig:blades_GP_imag_EX1b} and \ref{fig:blades_GP_real_EX1b} show that the differences in the natural frequencies of the blades resulted in significant variability in the data along the horizontal axis, for which the single GP is not well-suited. Because of this variation along the horizontal axis, the GP was forced to apply excessive damping when fitting the data. The NMSE was again calculated for the single GP ($ M = 13080 $),  and was unsurprisingly, much higher than those for the mixture, at 55.8 for the real fit and 65.1 for the imaginary fit. These results suggest that it is more appropriate to use a mixture of GPs in cases where the data have significant horizontal variability, as occurred here.

	\vspace{12pt} 
	\subsection{A form using an unsupervised mixture of Gaussian processes (OMGP)}

	An overlapping mixture of Gaussian processes (OMGP), assumes that the number of trajectory functions are known \emph{a priori}, but does not assume which data belong to a given class. In this case, because four helicopter blades contributed to the data set, it was assumed that four trajectories (i.e., $ K = 4 $ FRFs) describe the population form. The data were then classified and fitted by approximating the intractable posterior distribution $ p(\{\mathbf{f}^{(k)}\}_{k=1}^{K},\mathbf{Z} \;|\; \mathbf{x},\mathbf{y}) $ via the OMGP approach described in \cite{LAZAROGREDILLA20121386}, using the variational inference and EM scheme laid out in Eqs.\ (\ref{eq:latentfuncOMGP}-\ref{eq:PredictiveEqs3}).
	
\begin{figure}[h!]
		\centering
		\subfloat[\label{fig:blades_GP_real_EX2}]{\includegraphics[width=1\textwidth]{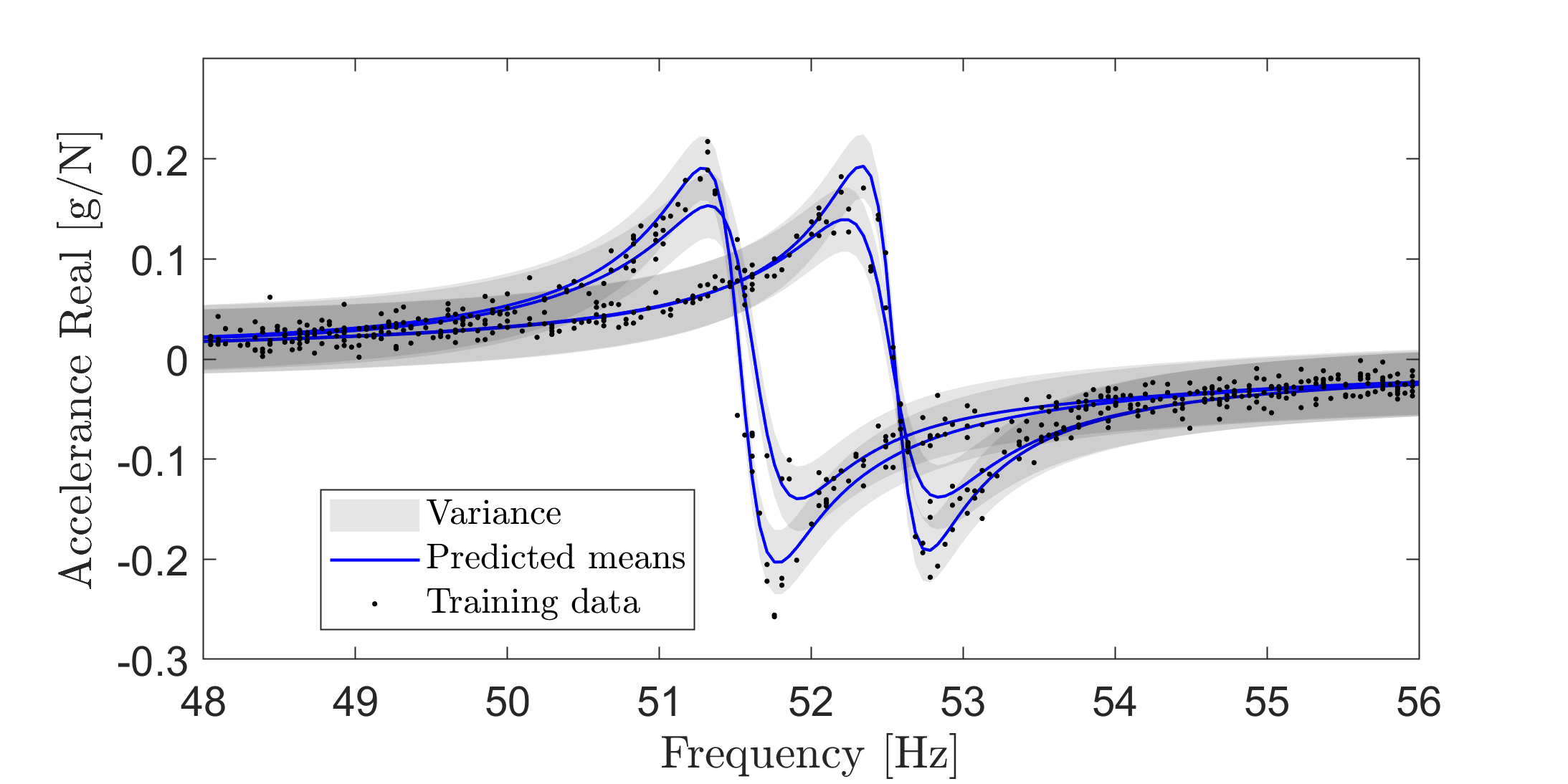}} \\
		\subfloat[\label{fig:blades_GP_imag_EX2}]{\includegraphics[width=1\textwidth]{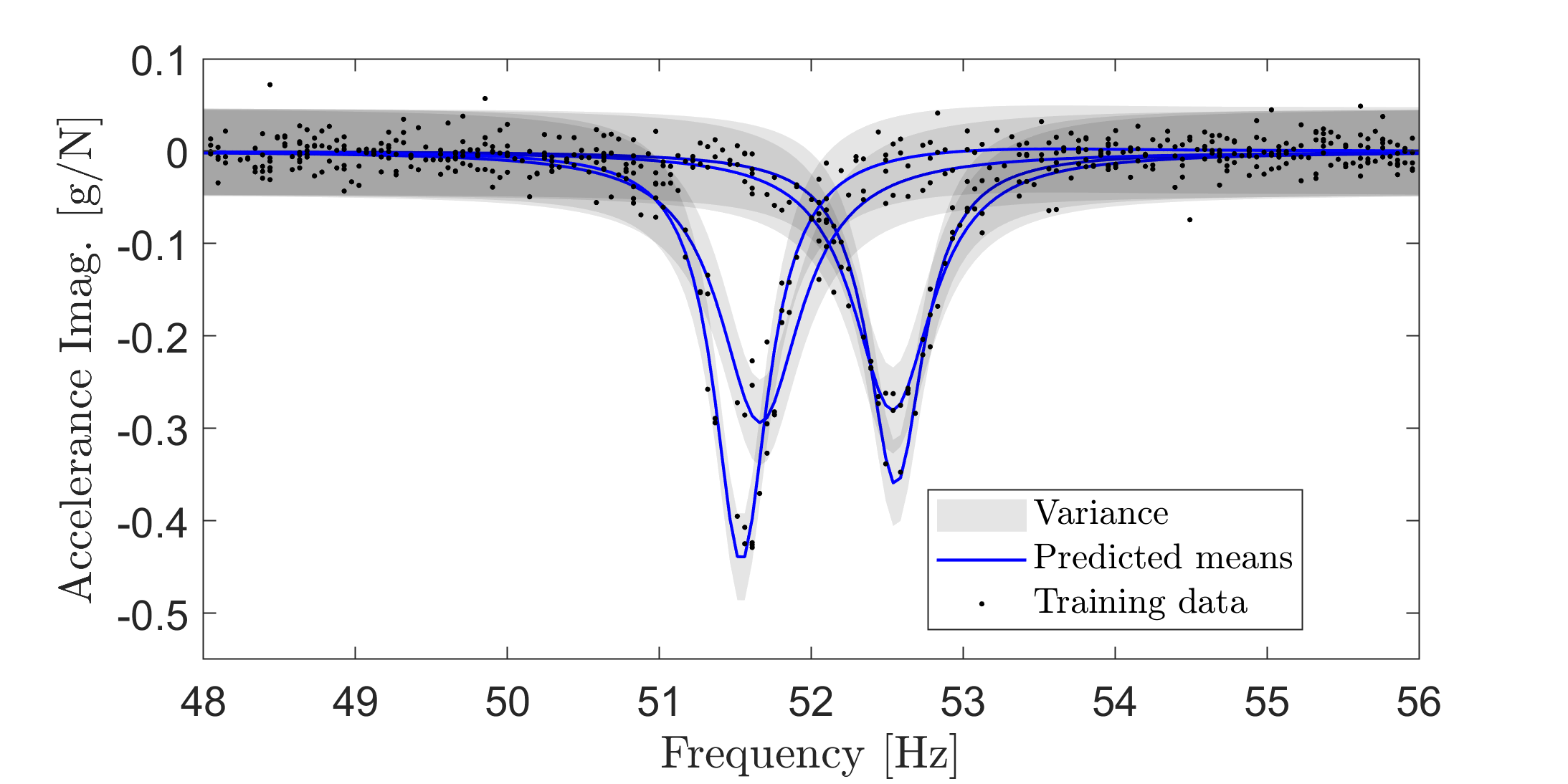}} \\ [-0.5ex]
		\caption{OMGP predictions using (a) real and (b) imaginary FRF data.}
	\end{figure}

	As with the previous example, a narrow frequency band containing the fifth bending mode of the blades, located between 48 and 56 Hz, was isolated for analysis and treated as an SDOF. Real and imaginary parts of the FRF were fit separately via Eqs.\ (\ref{eq:latentfuncOMGP}-\ref{eq:PredictiveEqs3}). The same replicated data with added noise were used as in the previous case, and 600 training points were randomly selected from these data. Likewise, the modal damping FRF model from Eq.\ (\ref{eq:modalFRF}) was used to generate a mean function, and modal damping, residue, and natural frequency were used as the mean-function hyperparameters. Upper and lower bounds were applied to constrain the hyperparameter optimisation, based on prior knowledge of the blades and visual inspection of the data, and the FRF direction was again assumed known. Because the real part of the FRF was more separable than the imaginary part (indeed, visual inspection of the FRFs showed that four distinct trajectories in the training data were harder to decipher from the imaginary part versus the real part), the OMGP performed much better for a variety of initialisations on the real data compared to the imaginary data. As such, the real data were fitted first, and an optimal solution was found by randomly initialising the process (i.e., via prior guesses for the hyperparameters), ten times while keeping the training set fixed, and choosing the solution with the maximum lower bound and therefore the highest marginal likelihood. The initialisation parameters were chosen randomly, but were restricted to reasonable intervals in the same manner as the hyperparameters were bounded during the optimisation. Once learnt, the optimised hyperparameters for the real data were used to initialise the OMGP for the imaginary data. The posterior predictive mean of the OMGP is plotted with the mean function, variance, and training data for the real and imaginary FRFs in Figures \ref{fig:blades_GP_real_EX2} and \ref{fig:blades_GP_imag_EX2}. 
	
	Visual inspection of Figures \ref{fig:blades_GP_real_EX2} and \ref{fig:blades_GP_imag_EX2} show that the OMGP provided comparable results to the supervised mixture from the previous example (shown in Figures \ref{fig:blades_GP_real_EX1a} and \ref{fig:blades_GP_imag_EX1a}). Likewise, the NMSE were again calculated via Eq.\ (\ref{eq:NMSE}) using an independent test data set ($ M = 13080 $). The calculated NMSE (2.3 for the real fit and 6.5 for the imaginary fit) were similar to those obtained with the supervised mixture (2.4 for the real fit, and 6.9 for the imaginary fit). The NMSE and MSD (computed via Eq.\ (\ref{eq:MSD})) for the OMGP are listed in Table \ref{tab:model_metrics}, along with those computed for the supervised mixture and single GP in the previous example.

	\begin{table}
		\centering
		\caption{Model performance metrics for (OM)GP forms.\label{tab:model_metrics}}
	\begin{tabular}{*{8}{l}}
			\hline 
			\multicolumn{2}{l}{} &  \multicolumn{2}{l}{Single GP} & \multicolumn{2}{l}{S. mixture} & \multicolumn{2}{l}{OMGP} \\
			\multicolumn{2}{l}{} & \multicolumn{1}{l}{real} & \multicolumn{1}{l}{imag} & \multicolumn{1}{l}{real} & \multicolumn{1}{l}{imag} & \multicolumn{1}{l}{real} & \multicolumn{1}{l}{imag} \\ 
			\hline
			\multicolumn{2}{l}{NMSE} & 55.76 & 65.05 & 2.40 & 6.92 & 2.32 & 6.49 \\
			\multicolumn{2}{l}{MSD} & 1.01 & 0.71 & 0.99 & 0.98 & 0.76 & 0.57 \\
			\hline
		\end{tabular}
	\end{table}



	The OMGP provides a unified form that generalises the normal condition for a population of structures, despite slight changes in boundary conditions and differences among the nominally-identical members. Unlike the conventional GP, that cannot accommodate large amounts of input-dependent noise, the OMGP (and the supervised mixture of GPs) can account for the horizontal variance that results from normal variations. The OMGP further generalises from the supervised mixture of GPs by allowing for an unlabelled training data set.

\vspace{12pt} 
\section{Novelty detection via the form}

	One practical implementation of the OMGP form is for damage identification, where unseen data can be compared against the posterior predictive distributions of the OMGP and evaluated for novelty. This section demonstrates how novelty detection could be accomplished, using a technique similar to that involving the MSD in \cite{Bull_1,Bull_2,WORDEN2000647}, but using the marginal likelihood from Eq.\ (\ref{eq:PredictiveEqs2}) as a novelty index. Because the real and imaginary parts of the FRFs were fitted separately, the sum of the (negative log) marginal likelihood from each fit provided the novelty index. A Monte Carlo approach (as in \cite{WORDEN2000647}) was used to determine an appropriate normal-condition threshold. 
	
	To evaluate the sensitivity of the form to changes in natural frequency, real and imaginary SDOF FRFs were synthesised using Eq.\ (\ref{eq:modalFRF}) to simulate stiffness reduction resulting from damage, which may present as a downward shift in natural frequency. FRFs were generated for each blade using modal information from the fixed-free tests (i.e., residues and modal damping obtained via hyperparameter optimisation from fitting the OMGP prior mean functions – representing the `true' values for these parameters) and natural frequency incrementally decreased by 0.5\% for a maximum reduction of 3.5\% (i.e., the natural frequency in the band of interest associated with each blade was reduced such that it was 0.5\%, 1\%, 1.5\%, 2\%, 2.5\%, 3\%, and 3.5\% lower in frequency than the true, undamaged value). In other words, the `damaged' FRFs were very similar to the undamaged experimental FRFs, except that they were slightly lower in frequency. The simulated FRFs were copied 1000 times and random Gaussian noise with a magnitude equal to 5\% of the absolute peak value of the FRF was added to each copy. The negative log-marginal-likelihood was then computed for each copy, with the novelty index equal to the negative log sum of the marginal likelihoods for the real and imaginary parts. Likewise, the experimental data from the fixed-free tests for each blade were copied 1000 times and random Gaussian noise with a magnitude equal to 5\% of the absolute peak value of the FRF was added to each copy. As with the simulated data, the copied experimental data (with added noise), were tested against the OMGP form, to evaluate the performance of the novelty detector for data known to be inlying.
	
	To identify a suitable normal-condition threshold, a similar technique to that applied in \cite{Bull_1,Bull_2,WORDEN2000647} was used, where 1000 samples were randomly selected from the normal-condition data (i.e., experimental FRFs, replicated with added noise) used to train the form. The negative log-marginal-likelihood was then calculated for the samples, for a large number of trials. The critical value was the threshold with a certain percentage of the calculated values below it. This work used a 99\% confidence interval, or $ 2.58\sigma $.
		
\begin{figure}[h!]
		\centering
		\subfloat[\label{fig:Synth1_2}]{\includegraphics[width=1\textwidth]{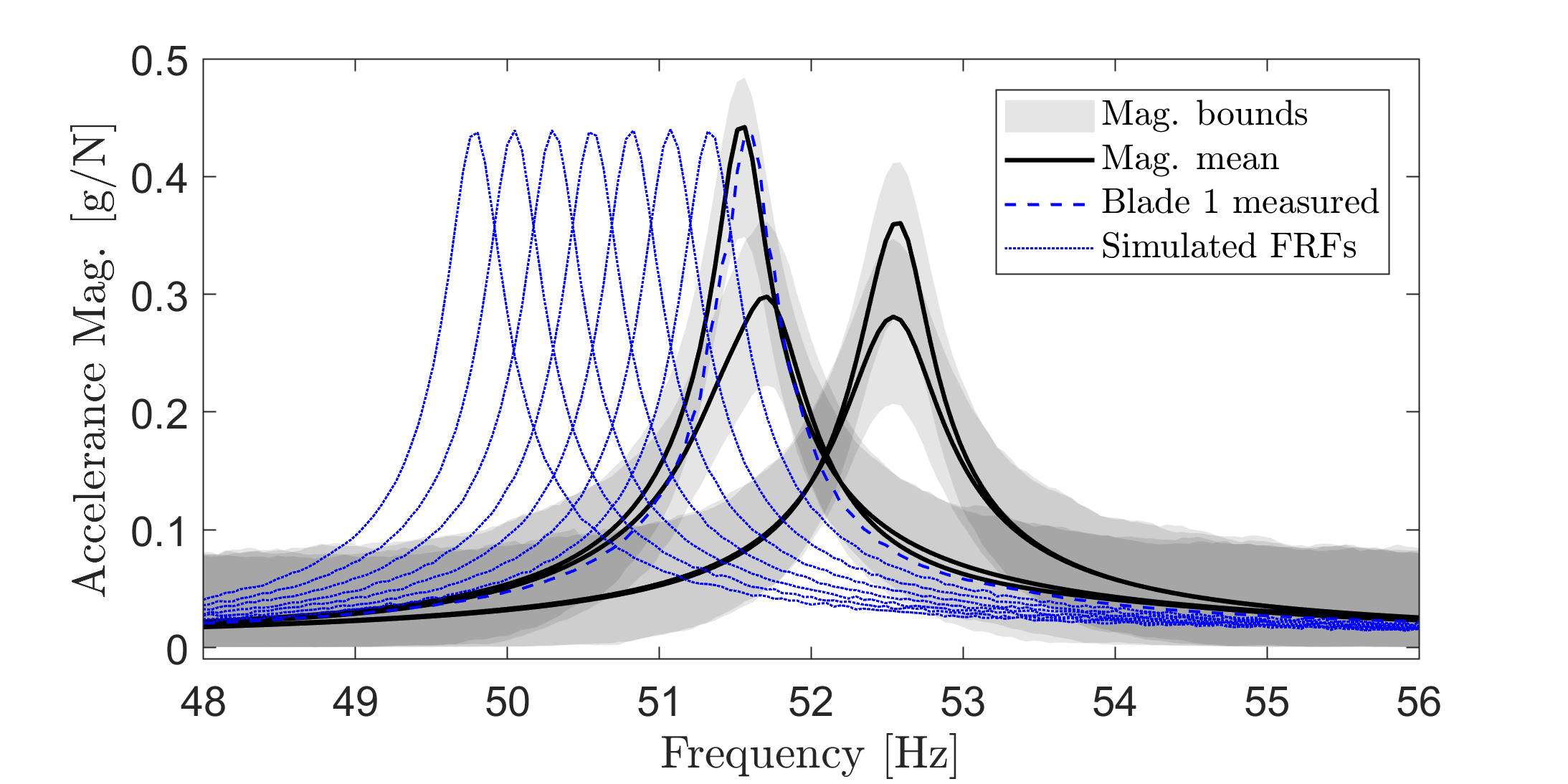}} \\  [-0.5ex]
		\subfloat[\label{fig:Synth2_2}]{\includegraphics[width=1\textwidth]{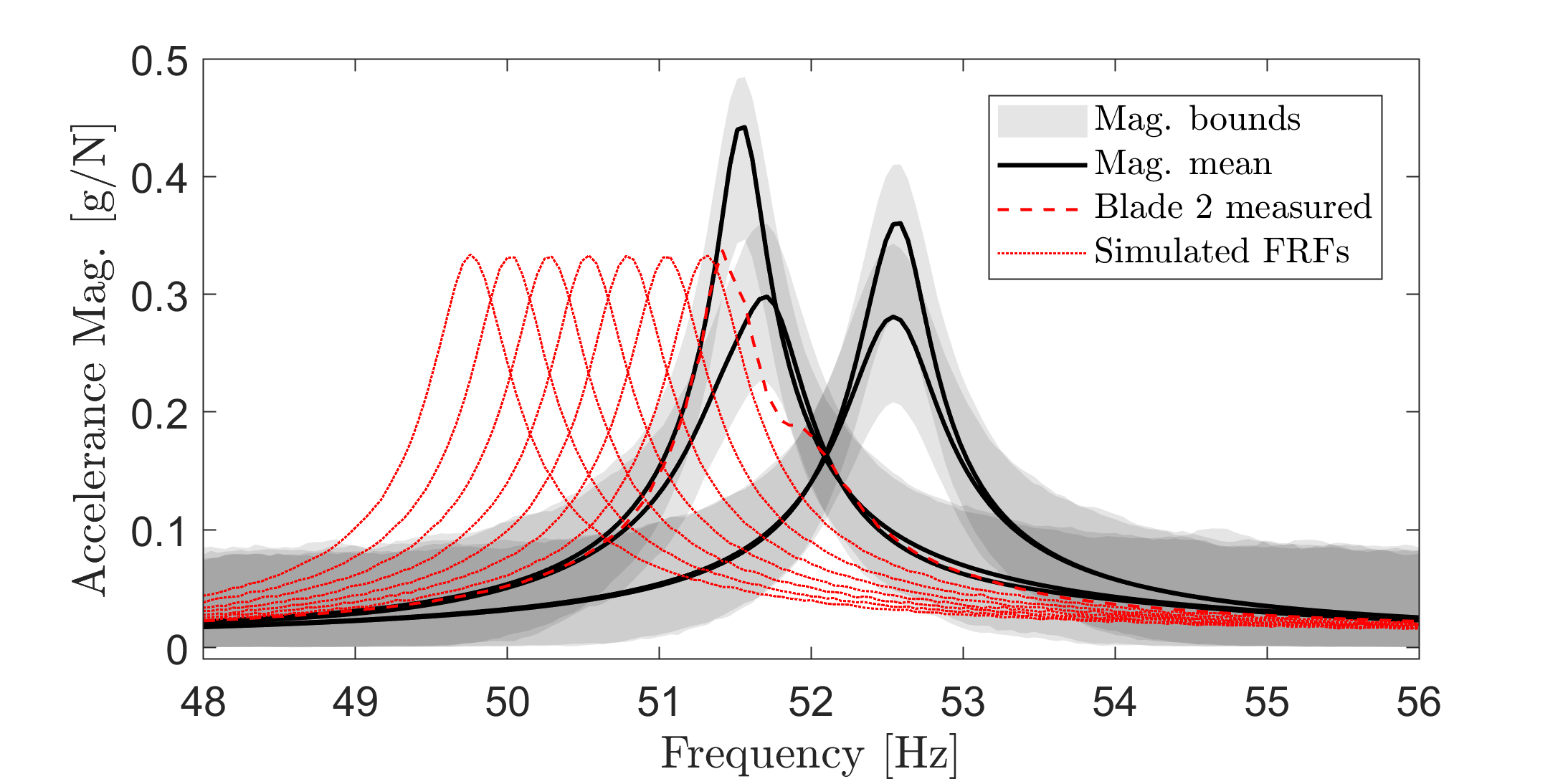}} \\
		\caption{Synthesised FRFs with incrementally decreasing natural frequency for (a) Blade 1 and (b) Blade 2, superimposed on the magnitude of the predicted OMGP.}
	\end{figure}

\begin{figure}[h!]
		\centering
		\subfloat[\label{fig:Synth3_2}]{\includegraphics[width=1\textwidth]{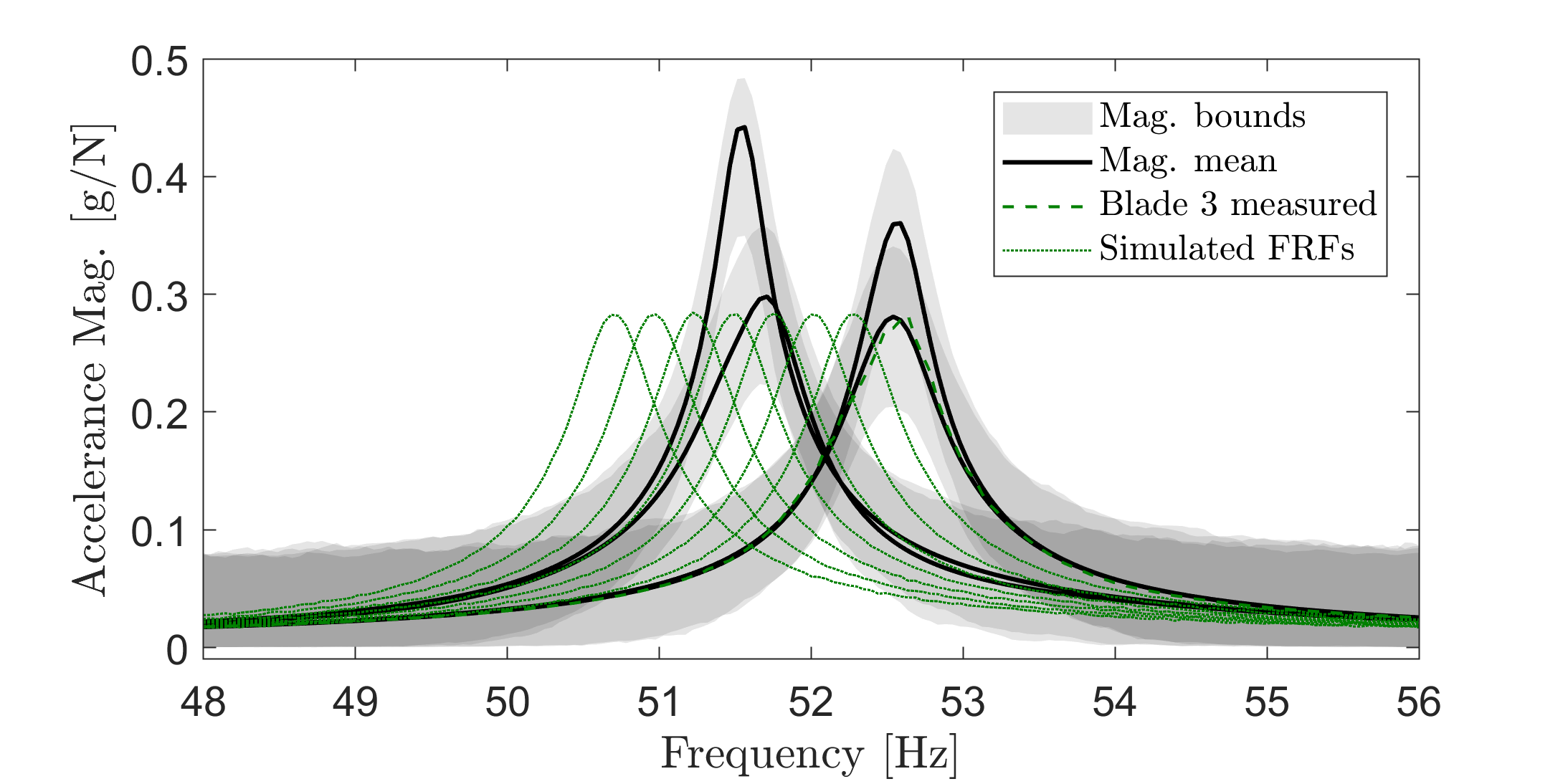}} \\  [-0.5ex]
		\subfloat[\label{fig:Synth4_2}]{\includegraphics[width=1\textwidth]{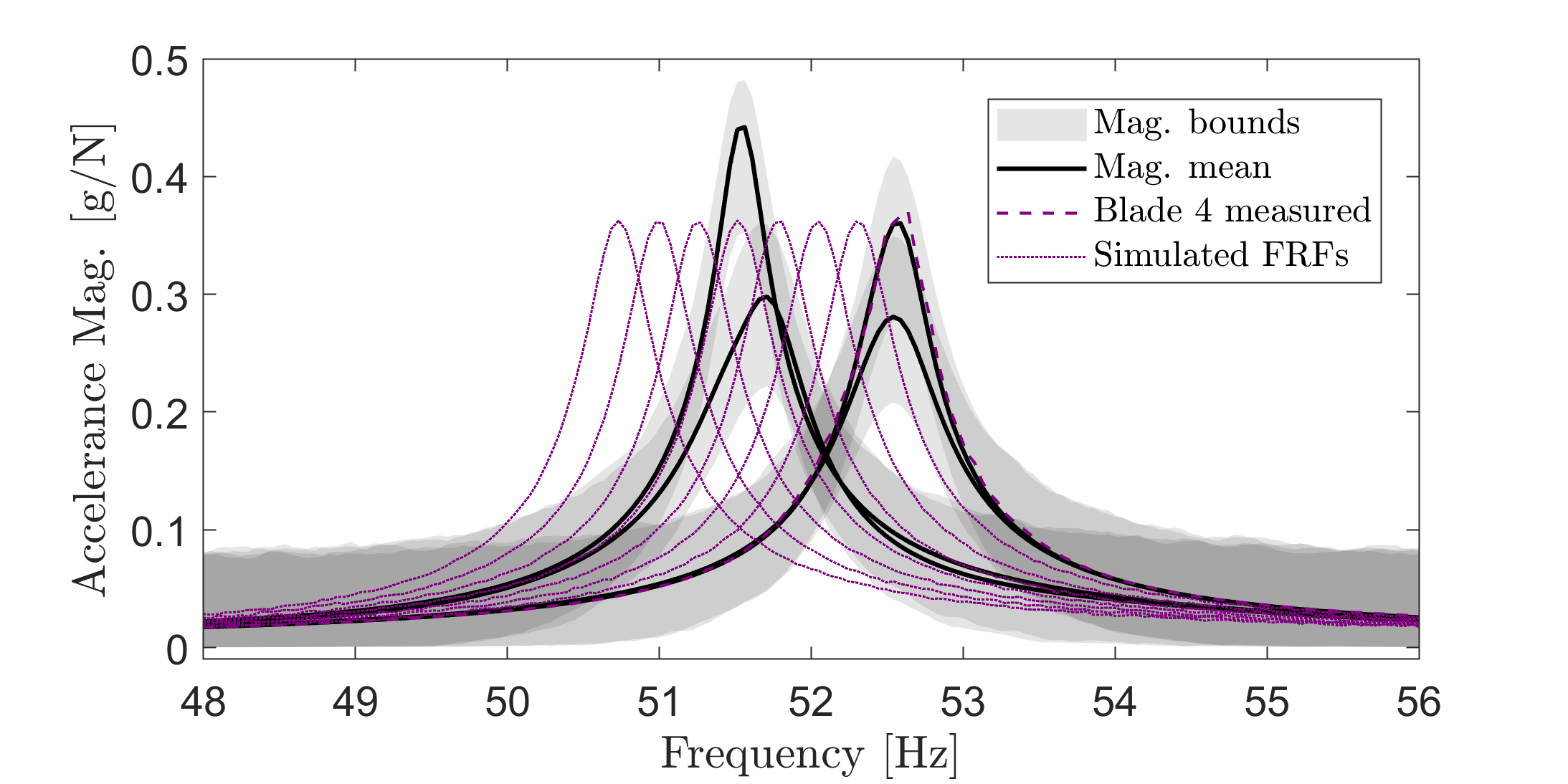}} \\
		\caption{Synthesised FRFs with incrementally decreasing natural frequency for (a) Blade 3 and (b) Blade 4, superimposed on the magnitude of the predicted OMGP.}
	\end{figure}

\begin{figure}[h!]
		\subfloat[\label{fig:OA1_2}]{\includegraphics[width=\textwidth]{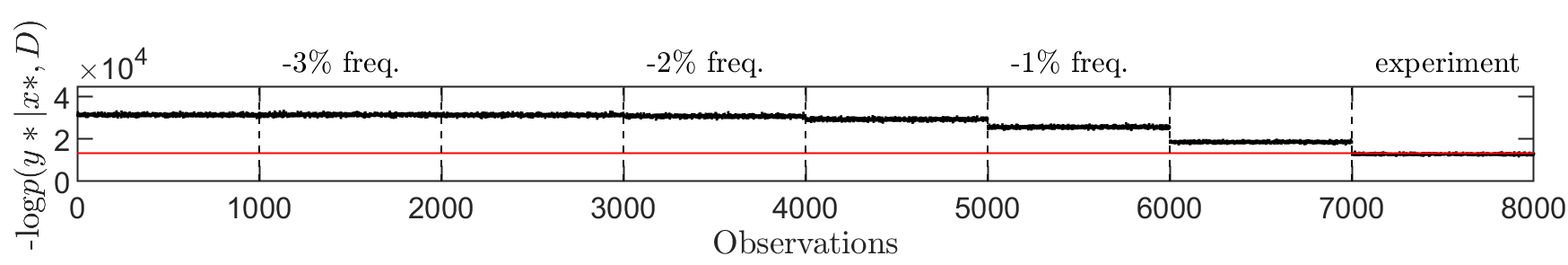}} \\[-0.25ex]
		\subfloat[\label{fig:OA2_2}]{\includegraphics[width=\textwidth,trim = {0cm 0cm 0cm 0.2cm},clip]{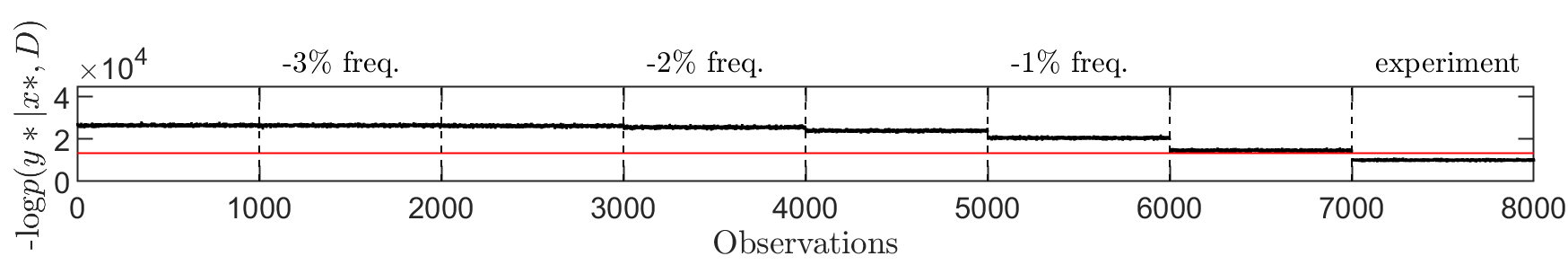}} \\[-0.25ex]
		\subfloat[\label{fig:OA3_2}]{\includegraphics[width=\textwidth,trim = {0cm 0cm 0cm 0.2cm},clip]{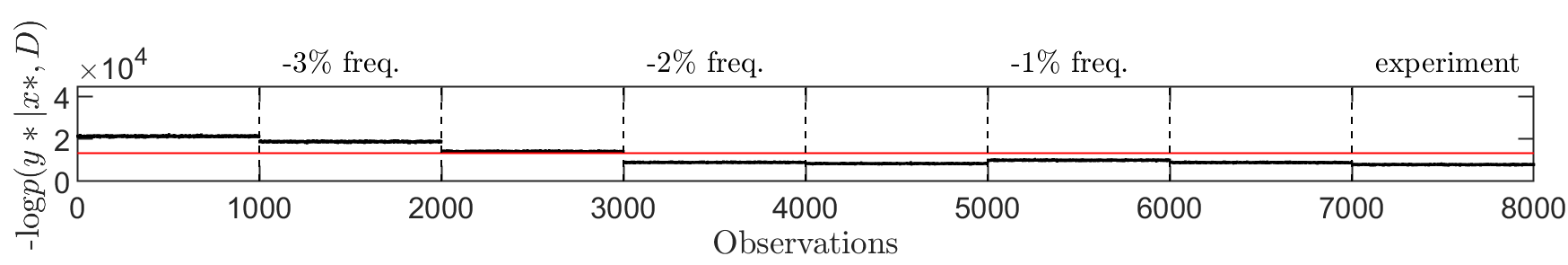}} \\[-0.25ex]
		\subfloat[\label{fig:OA4_2}]{\includegraphics[width=\textwidth,trim = {0cm 0cm 0cm 0.2cm},clip]{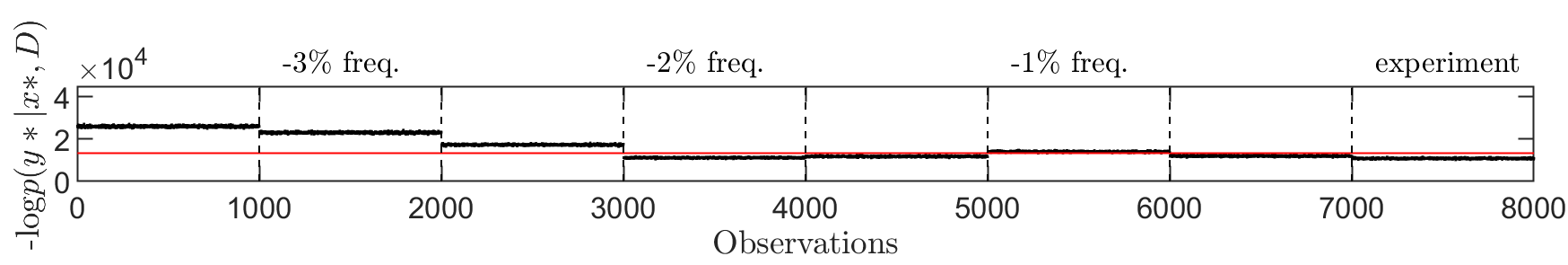}}
		\caption{Negative log-marginal-likelihood computed using experimental and simulated damaged results as test data for (a) Blade 1, (b) Blade 2, (c) Blade 3, and (d) \mbox{Blade 4}.}
	\end{figure}

	An approximation of the FRF form magnitude was computed for visualisation purposes. The posterior predictive distributions of the real and imaginary OMGPs were sampled 10,000 times, and the FRF magnitudes were computed from these samples. The mean and uncertainty bounds of the resulting magnitudes are shown in Figures \ref{fig:Synth1_2}-\ref{fig:Synth4_2}, along with the magnitudes of the simulated FRFs. Note that in Figures \ref{fig:Synth1_2}-\ref{fig:Synth4_2}, ‘Mag. mean’ refers to the mean of the FRF magnitudes as computed from samples from the OMGP posterior while ‘Mag. bounds’ shows the range within the magnitudes. The negative log-marginal-likelihood computed using the experimental data from the fixed-free tests as test data and the various synthesised FRFs as test data for Blades 1-4 are shown in Figures \ref{fig:OA1_2}-\ref{fig:OA4_2}, respectively.
	
	
	Examining the negative log-marginal-likelihood from right to left to correspond with the downward frequency shift of the FRFs, Figure \ref{fig:OA1_2} shows that -log$p(\mathbf{y}_*|\mathbf{x}_*,\mathcal{D})$ computed using replicated experimental FRFs for Blade 1 approached but did not surpass the threshold (shown in Figures \ref{fig:OA1_2}-\ref{fig:OA4_2} as a solid red line), which is expected, given that the peak belonging to Blade 1 (as shown in Figure \ref{fig:Synth1_2}) was located at the edge of the training data with respect to frequency. As the natural frequency of the synthesised FRFs decreased, the novelty index became increasingly outlying. In Figure \ref{fig:OA2_2}, similar behaviour for Blade 2 is visible, although the novelty index was further below the threshold for each case as the FRF from Blade 2 had a lower magnitude than that of Blade 1, as shown in Figures \ref{fig:Synth1_2} and \ref{fig:Synth2_2}. Figure \ref{fig:OA3_2} shows that for Blade 3, the novelty index gradually increased until approaching the threshold at -1.5\%, which corresponded to the space between the peak groupings. As the natural frequency of the simulated FRFs for Blade 3 decreased further (-1.5\% to -2\%), the novelty index fell, because the simulated FRFs aligned with the natural frequencies of Blades 1 and 2. As the natural frequency of the synthesised FRFs continued to decrease (-2.5\% to -3.5\%), the novelty index became increasingly outlying. Figure \ref{fig:OA4_2} shows similar results for Blade 4; however, because Blade 4 had a larger FRF magnitude than Blade 3, the novelty index was closer to the threshold. These results suggest that for Blades 1 and to a lesser extent Blade 3, the technique would be quite sensitive to downward frequency shifts. The technique would be less sensitive to frequency reductions for Blades 3 and 4, as a decrease in frequency may result in overlap with the normal-condition peaks for Blades 1 and 2. Note that this metric was computed in a functional sense, meaning that the test data included the full (band-limited) FRF, rather than individual test points, and the full covariance matrices of the OMGP were used to compute the novelty index. Considering the metric in a functional sense means that if the test data fit a function significantly different than the form (e.g., the test data were not an FRF), then the data could be flagged as outlying regardless of whether the majority of individual test points were within the variance bounds.


\vspace{12pt} 
\section{Concluding remarks}
	This paper presented data obtained from a comprehensive testing campaign on four healthy, nominally-identical helicopter blades, and used the measured data to develop a generic model, called a \emph{form}, to represent the small `population.' In the first example presented, a supervised mixture of GPs was used to develop the form, where the labels were assumed known \emph{a priori}. The second example used an (unsupervised) OMGP to develop the form, where the data classes were presumed unknown and needed to be inferred via the OMGP (variational inference/EM scheme) approach. Comparable results were obtained from both techniques. In addition, the posterior predictive distributions from the OMGP were applied to a novelty detection approach to evaluate new (simulated) data, providing insight into the generalisation capabilities of the model. 

	Population-based SHM seeks to transfer valuable information across similar structures, such as normal conditions and damaged states. The results presented herein demonstrate preliminary techniques for establishing a normal condition for a population of structures that are nominally-identical, but distinct, with consideration for variations such as discrepancies in material properties or fluctuations in boundary conditions. Although the proposed model provides an accurate representation of the population data, it does not (in its current form) consider correlation between the related functions. Further work will consider methods of inferring correlated task parameters for knowledge transfer within the group.

\vspace{12pt} 
\section{Acknowledgements}
The authors gratefully acknowledge the support of the UK Engineering and Physical Sciences Research Council (EPSRC), via grant references EP/R003645\-/1 and EP/R004900/1. L.A. Bull was supported by Wave 1 of the UKRI Strategic Priorities Fund under the EPSRC grant EP/W006022/1, particularly the \textit{Ecosystems of Digital Twins} theme within that grant and the Alan Turing Institute. For the purpose of open access, the author(s) has/have applied a Creative Commons Attribution (CC BY) licence to any Author Accepted Manuscript version arising.

This research made use of The Laboratory for Verification and Validation (LVV), which was funded by the EPSRC (grant numbers EP/J013714/1 and EP/N010884/1), the European Regional Development Fund (ERDF) and the University of Sheffield. The authors would like to extend special thanks to Michael Dutchman, for assisting with experimental setup. \\




\bibliographystyle{elsarticle-num-names} 
\bibliography{references}

\begin{thebibliography}{27}
\expandafter\ifx\csname natexlab\endcsname\relax\def\natexlab#1{#1}\fi
\providecommand{\url}[1]{\texttt{#1}}
\providecommand{\href}[2]{#2}
\providecommand{\path}[1]{#1}
\providecommand{\DOIprefix}{doi:}
\providecommand{\ArXivprefix}{arXiv:}
\providecommand{\URLprefix}{URL: }
\providecommand{\Pubmedprefix}{pmid:}
\providecommand{\doi}[1]{\href{http://dx.doi.org/#1}{\path{#1}}}
\providecommand{\Pubmed}[1]{\href{pmid:#1}{\path{#1}}}
\providecommand{\bibinfo}[2]{#2}
\ifx\xfnm\relax \def\xfnm[#1]{\unskip,\space#1}\fi
\bibitem[{Dowsett et~al.(2018)Dowsett, O’Boy, Walsh, Abolfathi, and
  Fisher}]{Dowsett}
\bibinfo{author}{A.~Dowsett}, \bibinfo{author}{D.~O’Boy},
  \bibinfo{author}{S.~Walsh}, \bibinfo{author}{A.~Abolfathi},
  \bibinfo{author}{S.~Fisher},
\newblock \bibinfo{title}{The prediction of measurement variability in an
  automotive application by the use of a coherence formulation},
\newblock \bibinfo{journal}{Proc. Inst. Mech. Eng. D: J. Automob. Eng.}
  \bibinfo{volume}{232} (\bibinfo{year}{2018}) \bibinfo{pages}{1694--1700}.
  \DOIprefix\doi{10.1177/0954407017734768}.
\bibitem[{Chen et~al.(2006)Chen, Duhamel, and Soize}]{CHEN200664}
\bibinfo{author}{C.~Chen}, \bibinfo{author}{D.~Duhamel},
  \bibinfo{author}{C.~Soize},
\newblock \bibinfo{title}{Probabilistic approach for model and data
  uncertainties and its experimental identification in structural dynamics:
  {Case} of composite sandwich panels},
\newblock \bibinfo{journal}{J. Sound Vib.} \bibinfo{volume}{294}
  (\bibinfo{year}{2006}) \bibinfo{pages}{64--81}.
  \DOIprefix\doi{10.1016/j.jsv.2005.10.013}.
\bibitem[{{L.A.\ Bull} et~al.(2021){L.A.\ Bull}, {P.A.\ Gardner}, Gosliga,
  {T.J.\ Rogers}, Dervilis, {E.J.\ Cross}, Papatheou, {A.E.\ Maquire}, Campos,
  and Worden}]{Bull_1}
\bibinfo{author}{{L.A.\ Bull}}, \bibinfo{author}{{P.A.\ Gardner}},
  \bibinfo{author}{J.~Gosliga}, \bibinfo{author}{{T.J.\ Rogers}},
  \bibinfo{author}{N.~Dervilis}, \bibinfo{author}{{E.J.\ Cross}},
  \bibinfo{author}{E.~Papatheou}, \bibinfo{author}{{A.E.\ Maquire}},
  \bibinfo{author}{C.~Campos}, \bibinfo{author}{K.~Worden},
\newblock \bibinfo{title}{Foundations of population-based {SHM}, {Part I}:
  {Homogeneous} populations and forms},
\newblock \bibinfo{journal}{Mech. Syst. Signal Process.} \bibinfo{volume}{148}
  (\bibinfo{year}{2021}) \bibinfo{pages}{107141}.
  \DOIprefix\doi{10.1016/j.ymssp.2020.107141}.
\bibitem[{{L.A.\ Bull} et~al.(2019){L.A.\ Bull}, {T.J.\ Rogers}, Dervilis,
  {E.J.\ Cross}, and Worden}]{Bull_2}
\bibinfo{author}{{L.A.\ Bull}}, \bibinfo{author}{{T.J.\ Rogers}},
  \bibinfo{author}{N.~Dervilis}, \bibinfo{author}{{E.J.\ Cross}},
  \bibinfo{author}{K.~Worden},
\newblock \bibinfo{title}{A {Gaussian} process form for population-based
  structural health monitoring},
\newblock in: \bibinfo{booktitle}{Proceedings of the 13th International
  Conference on Damage Assessment of Structures (DAMAS 2019), Porto, Portugal},
  \bibinfo{organization}{Springer}, \bibinfo{year}{July 9-10, 2019}.
\bibitem[{Gosliga et~al.(2021)Gosliga, {P.A.\ Gardner}, {L.A.\ Bull}, Dervilis,
  and Worden}]{gosliga2021foundations}
\bibinfo{author}{J.~Gosliga}, \bibinfo{author}{{P.A.\ Gardner}},
  \bibinfo{author}{{L.A.\ Bull}}, \bibinfo{author}{N.~Dervilis},
  \bibinfo{author}{K.~Worden},
\newblock \bibinfo{title}{Foundations of population-based {SHM}, {Part II}:
  Heterogeneous populations--graphs, networks, and communities},
\newblock \bibinfo{journal}{Mech. Syst. Signal Process.} \bibinfo{volume}{148}
  (\bibinfo{year}{2021}) \bibinfo{pages}{107144}.
  \DOIprefix\doi{10.1016/j.ymssp.2020.107144}.
\bibitem[{{P.A.\ Gardner} et~al.(2021){P.A.\ Gardner}, {L.A.\ Bull}, Gosliga,
  Dervilis, and Worden}]{gardner2021foundations}
\bibinfo{author}{{P.A.\ Gardner}}, \bibinfo{author}{{L.A.\ Bull}},
  \bibinfo{author}{J.~Gosliga}, \bibinfo{author}{N.~Dervilis},
  \bibinfo{author}{K.~Worden},
\newblock \bibinfo{title}{Foundations of population-based {SHM}, {Part III}:
  Heterogeneous populations--mapping and transfer},
\newblock \bibinfo{journal}{Mech. Syst. Signal Process.} \bibinfo{volume}{149}
  (\bibinfo{year}{2021}) \bibinfo{pages}{107142}.
  \DOIprefix\doi{10.1016/j.ymssp.2020.107142}.
\bibitem[{Worden et~al.(2002)Worden, Sohn, and Farrar}]{Worden2002NoveltyDI}
\bibinfo{author}{K.~Worden}, \bibinfo{author}{H.~Sohn},
  \bibinfo{author}{C.~Farrar},
\newblock \bibinfo{title}{Novelty detection in a changing environment:
  {Regression} and interpolation approaches},
\newblock \bibinfo{journal}{J. Sound Vib.} \bibinfo{volume}{258}
  (\bibinfo{year}{2002}) \bibinfo{pages}{741--761}.
  \DOIprefix\doi{10.1006/jsvi.2002.5148}.
\bibitem[{Alampalli(1998)}]{Alampalli}
\bibinfo{author}{S.~Alampalli},
\newblock \bibinfo{title}{Influence of in-service environment on modal
  parameters},
\newblock in: \bibinfo{booktitle}{Proceedings of the 16th International Modal
  Analysis Conference (IMAC)}, \bibinfo{publisher}{Society for Experimental
  Mechanics, Santa Barbara, CA, USA}, \bibinfo{year}{1998}.
\bibitem[{Cawley(1997)}]{Cawley}
\bibinfo{author}{P.~Cawley},
\newblock \bibinfo{title}{Long range inspection of structures using low
  frequency ultrasound},
\newblock in: \bibinfo{booktitle}{Proceedings of the International Conference
  on Damage Assessment of Structures (DAMAS 97), Sheffield, UK},
  \bibinfo{organization}{Sheffield Academic Press}, \bibinfo{year}{June 30 -
  July 2, 1997}, p. \bibinfo{pages}{1–17}.
\bibitem[{Sohn(2007)}]{HoonSohn}
\bibinfo{author}{H.~Sohn},
\newblock \bibinfo{title}{Effects of environmental and operational variability
  on structural health monitoring},
\newblock \bibinfo{journal}{Philos. Trans. A: Math. Phys. Eng. Sci.}
  \bibinfo{volume}{365} (\bibinfo{year}{2007}) \bibinfo{pages}{539--560}.
  \DOIprefix\doi{10.1098/rsta.2006.1935}.
\bibitem[{{L.A.\ Bull} et~al.(2021){L.A.\ Bull}, {P.A.\ Gardner}, {T.J.\
  Rogers}, Dervilis, {E.J.\ Cross}, Papatheou, {A.E.\ Maquire}, Campos, and
  Worden}]{Bull_3}
\bibinfo{author}{{L.A.\ Bull}}, \bibinfo{author}{{P.A.\ Gardner}},
  \bibinfo{author}{{T.J.\ Rogers}}, \bibinfo{author}{N.~Dervilis},
  \bibinfo{author}{{E.J.\ Cross}}, \bibinfo{author}{E.~Papatheou},
  \bibinfo{author}{{A.E.\ Maquire}}, \bibinfo{author}{C.~Campos},
  \bibinfo{author}{K.~Worden},
\newblock \bibinfo{title}{Bayesian modelling of multivalued power curves from
  an operational wind farm},
\newblock \bibinfo{journal}{Mech. Syst. Signal Process.}
  (\bibinfo{year}{2021}) \bibinfo{pages}{108530}.
  \DOIprefix\doi{10.1016/j.ymssp.2021.108530}.
\bibitem[{Worden et~al.(2000)Worden, Manson, and Fieller}]{WORDEN2000647}
\bibinfo{author}{K.~Worden}, \bibinfo{author}{G.~Manson},
  \bibinfo{author}{N.~Fieller},
\newblock \bibinfo{title}{Damage detection using outlier analysis},
\newblock \bibinfo{journal}{J. Sound Vib.} \bibinfo{volume}{229}
  (\bibinfo{year}{2000}) \bibinfo{pages}{647--667}.
  \DOIprefix\doi{10.1006/JSVI.1999.2514}.
\bibitem[{Lázaro-Gredilla et~al.(2012)Lázaro-Gredilla, {Van Vaerenbergh}, and
  {N.D.\ Lawrence}}]{LAZAROGREDILLA20121386}
\bibinfo{author}{M.~Lázaro-Gredilla}, \bibinfo{author}{S.~{Van Vaerenbergh}},
  \bibinfo{author}{{N.D.\ Lawrence}},
\newblock \bibinfo{title}{Overlapping mixtures of {Gaussian} processes for the
  data association problem},
\newblock \bibinfo{journal}{Pattern Recognit.} \bibinfo{volume}{45}
  (\bibinfo{year}{2012}) \bibinfo{pages}{1386--1395}.
  \DOIprefix\doi{10.1016/j.patcog.2011.10.004}.
\bibitem[{Hong et~al.(2011)Hong, {B.I.\ Epureanu}, {M.P.\ Castanier}, and
  {D.J.\ Gorsich}}]{HONG20111091}
\bibinfo{author}{S.-K. Hong}, \bibinfo{author}{{B.I.\ Epureanu}},
  \bibinfo{author}{{M.P.\ Castanier}}, \bibinfo{author}{{D.J.\ Gorsich}},
\newblock \bibinfo{title}{Parametric reduced-order models for predicting the
  vibration response of complex structures with component damage and
  uncertainties},
\newblock \bibinfo{journal}{J. Sound Vib.} \bibinfo{volume}{330}
  (\bibinfo{year}{2011}) \bibinfo{pages}{1091--1110}.
  \DOIprefix\doi{10.1016/j.jsv.2010.09.022}.
\bibitem[{Yang and Kessissoglou(2013)}]{Yang}
\bibinfo{author}{J.~Yang}, \bibinfo{author}{N.~Kessissoglou},
\newblock \bibinfo{title}{Modal analysis of structures with uncertainties using
  polynomial chaos expansion},
\newblock in: \bibinfo{booktitle}{Proceedings of Acoustics},
  \bibinfo{organization}{Australian Acoustical Society, Victor Harbor,
  Australia}, \bibinfo{year}{Nov. 17-20, 2013}.
\bibitem[{{E.J.\ Cross} et~al.(2022){E.J.\ Cross}, {S.J.\ Gibson}, {M.R.\
  Jones}, {D.J.\ Pitchforth}, Zhang, and {T.J.\ Rogers}}]{Cross2022}
\bibinfo{author}{{E.J.\ Cross}}, \bibinfo{author}{{S.J.\ Gibson}},
  \bibinfo{author}{{M.R.\ Jones}}, \bibinfo{author}{{D.J.\ Pitchforth}},
  \bibinfo{author}{S.~Zhang}, \bibinfo{author}{{T.J.\ Rogers}},
  \bibinfo{title}{Physics-Informed Machine Learning for Structural Health
  Monitoring}, \bibinfo{publisher}{Springer International Publishing},
  \bibinfo{year}{2022}, pp. \bibinfo{pages}{347--367}.
  \DOIprefix\doi{{10.1007/978-3-030-81716-9}}.
\bibitem[{Worden and {G.R.\ Tomlinson}(2001)}]{wordennonlinearity}
\bibinfo{author}{K.~Worden}, \bibinfo{author}{{G.R.\ Tomlinson}},
  \bibinfo{title}{Nonlinearity in Structural Dynamics: {Detection},
  Identification and Modelling}, \bibinfo{publisher}{Institute of Physics
  Publishing, Bristol, UK}, \bibinfo{year}{2001}.
\bibitem[{{K.P.\ Murphy}(2012)}]{murphy2012machine}
\bibinfo{author}{{K.P.\ Murphy}}, \bibinfo{title}{Machine Learning: A
  Probabilistic Perspective}, \bibinfo{publisher}{The MIT Press, Cambridge, MA,
  USA}, \bibinfo{year}{2012}.
\bibitem[{{C.E.\ Rasmussen} and {C.K.I.\ Williams}(2006)}]{3569}
\bibinfo{author}{{C.E.\ Rasmussen}}, \bibinfo{author}{{C.K.I.\ Williams}},
  \bibinfo{title}{{Gaussian} Processes for Machine Learning}, Adaptive
  Computation and Machine Learning, \bibinfo{publisher}{The MIT Press},
  \bibinfo{address}{Cambridge, MA, USA}, \bibinfo{year}{2006}.
\bibitem[{{M.C.K.\ Tay} and Laugier(2007)}]{tay2008modelling}
\bibinfo{author}{{M.C.K.\ Tay}}, \bibinfo{author}{C.~Laugier},
\newblock \bibinfo{title}{Modelling smooth paths using {Gaussian} processes},
\newblock in: \bibinfo{booktitle}{Proceedings of the International Conference
  on Field and Service Robotics (FSR07)}, \bibinfo{organization}{Springer,
  Chamonix, France}, \bibinfo{year}{July 9-12, 2007}, pp.
  \bibinfo{pages}{381--390}.
\bibitem[{{D.M.\ Blei} et~al.(2017){D.M.\ Blei}, Kucukelbir, and {J.D.\
  McAuliffe}}]{VIreview}
\bibinfo{author}{{D.M.\ Blei}}, \bibinfo{author}{A.~Kucukelbir},
  \bibinfo{author}{{J.D.\ McAuliffe}},
\newblock \bibinfo{title}{Variational inference: A review for statisticians},
\newblock \bibinfo{journal}{J. Am. Stat. Assoc.} \bibinfo{volume}{112}
  (\bibinfo{year}{2017}) \bibinfo{pages}{859–877}.
\bibitem[{{D.J.C.\ MacKay}(2003)}]{mackay2003information}
\bibinfo{author}{{D.J.C.\ MacKay}}, \bibinfo{title}{Information Theory,
  Inference and Learning Algorithms}, \bibinfo{publisher}{Cambridge University
  Press, Cambridge, UK}, \bibinfo{year}{2003}.
\bibitem[{{N.J.\ King} and {N.D.\ Lawrence}(2006)}]{king2006fast}
\bibinfo{author}{{N.J.\ King}}, \bibinfo{author}{{N.D.\ Lawrence}},
\newblock \bibinfo{title}{Fast variational inference for {Gaussian} process
  models through {KL}-correction},
\newblock in: \bibinfo{booktitle}{Proceedings of the 17th European Conference
  on Machine Learning (EMCL 06)}, \bibinfo{organization}{Springer-Verlag,
  Berlin, Germany}, \bibinfo{year}{Sept. 18-22, 2006}, pp.
  \bibinfo{pages}{270--281}.
\bibitem[{L{\'a}zaro-Gredilla and Titsias(2011)}]{lazaro2011variational}
\bibinfo{author}{M.~L{\'a}zaro-Gredilla}, \bibinfo{author}{M.~K. Titsias},
\newblock \bibinfo{title}{Variational heteroscedastic {Gaussian} process
  regression},
\newblock in: \bibinfo{booktitle}{Proceedings of the 28th International
  Conference on Machine Learning (ICML 11)}, \bibinfo{organization}{Omnipress,
  Bellevue, WA, USA}, \bibinfo{year}{June 28 - July 2, 2011}.
\bibitem[{{J.S.\ Bendat} and {A.G.\ Piersol}(2011)}]{bendat2011random}
\bibinfo{author}{{J.S.\ Bendat}}, \bibinfo{author}{{A.G.\ Piersol}},
  \bibinfo{title}{Random Data: {A}nalysis and Measurement Procedures (4th
  ed.)}, \bibinfo{publisher}{John Wiley \& Sons, New York, NY},
  \bibinfo{year}{2011}.
\bibitem[{{T.A.\ Dardeno} et~al.(2022{\natexlab{a}}){T.A.\ Dardeno}, {L.A.\
  Bull}, Dervilis, and Worden}]{DardenoIWSHM1}
\bibinfo{author}{{T.A.\ Dardeno}}, \bibinfo{author}{{L.A.\ Bull}},
  \bibinfo{author}{N.~Dervilis}, \bibinfo{author}{K.~Worden},
\newblock \bibinfo{title}{Investigating experimental repeatability and feature
  consistency in vibration-based {SHM}},
\newblock in: \bibinfo{booktitle}{Proceedings of the 13th International
  Workshop on Structural Health Monitoring ({IWSHM})},
  \bibinfo{year}{2022}{\natexlab{a}}.
\bibitem[{{T.A.\ Dardeno} et~al.(2022{\natexlab{b}}){T.A.\ Dardeno},
  Haywood-Alexander, {R.S.\ Mills}, {L.A.\ Bull}, Dervilis, and
  Worden}]{DardenoIWSHM2}
\bibinfo{author}{{T.A.\ Dardeno}}, \bibinfo{author}{M.~Haywood-Alexander},
  \bibinfo{author}{{R.S.\ Mills}}, \bibinfo{author}{{L.A.\ Bull}},
  \bibinfo{author}{N.~Dervilis}, \bibinfo{author}{K.~Worden},
\newblock \bibinfo{title}{Investigating the effects of ambient temperature on
  feature consistency in vibration-based {SHM}},
\newblock in: \bibinfo{booktitle}{Proceedings of the 13th International
  Workshop on Structural Health Monitoring ({IWSHM})},
  \bibinfo{year}{2022}{\natexlab{b}}.

\end{thebibliography}

\end{document}